%% file: arxiv.tex
\documentclass{article}

    \usepackage[preprint]{neurips_2021}

\usepackage{times}
\usepackage{epsfig}
\usepackage{graphicx}
\usepackage{amsmath}
\usepackage{dirtytalk}

\usepackage{microtype}
\usepackage{booktabs} 
\usepackage{amsfonts}       
\usepackage{nicefrac}       
\usepackage{microtype}      
\usepackage{amsthm}
\usepackage{enumitem}
\input{math.tex}

\newtheorem{theorem}{Theorem}
\newtheorem{corollary}{Corollary}[theorem]
\newtheorem{lemma}[theorem]{Lemma}

\newtheorem{innercustomthm}{Theorem}
\newenvironment{customthm}[1]
  {\renewcommand\theinnercustomthm{#1}\innercustomthm}
  {\endinnercustomthm}
\usepackage{enumitem}
\usepackage{algorithm}
\usepackage{algorithmic}
\usepackage{amssymb}
\usepackage{comment}
\usepackage{subcaption}
\usepackage{hyperref}

\newcommand{\nickins}[1]{\textcolor{black}{#1}}
\newcommand{\latestEdits}[1]{\textcolor{black}{#1}}
\usepackage[textsize=footnotesize]{todonotes}

\title{Iterative Averaging in the Quest for Best Test Error}

\author{%
  Diego Granziol\thanks{granziol.me} \\
  AI Theory\\
  Huawei \\
  \texttt{diego@robots.ox.ac.uk} \\
   \And
   Nicholas Baskerville \\
   Bristol University \\
   \texttt{n.p.baskerville@bristol.ac.uk} \\
   \AND
   Xingchen Wan \\
   Oxford University \\
   \texttt{xingchen.wan@robots.ox.ac.uk} \\
   \And
   Samuel Albanie \\
   Oxford University \\
   \texttt{samuel.albanie@robots.ox.ac.uk} \\
   \And
   Stephen Roberts \\
   Oxford University \\
   \texttt{sjrob@robots.ox.ac.uk} \\
}

\begin{document}

\maketitle

\begin{abstract}
We analyse and explain the increased generalisation performance \latestEdits{of} Iterate Averaging using a Gaussian Process perturbation model between the true and batch risk surface on the high dimensional quadratic.
We derive three phenomena \latestEdits{from our theoretical results:} (1) The importance of combining iterate averaging with large learning rates and regularisation for improved regularisation (2) Justification  for less frequent averaging. (3) That we expect adaptive gradient methods to work equally well or better with iterate averaging than their non adaptive counterparts. Inspired by these results\latestEdits{, together with} empirical investigations of the importance of appropriate regularisation for the solution diversity of the iterates, we propose two adaptive algorithms with iterate averaging.
\latestEdits{These} give significantly better results than SGD, require less tuning and do not require early stopping or validation set monitoring. We showcase the efficacy of our approach on the CIFAR-10/100, ImageNet and Penn Treebank datasets on a variety of modern and classical network architectures. 
\end{abstract}

\section{Introduction}
\label{sec:intro}

Deep Neural Network (DNN) models achieve state of the art performance in a plethora of problems, such as speech recognition, visual object image recognition, object detection, drug discovery and genomics \cite{lecun2015deep}. Of key interest is their ability to ``generalise'' to unseen data, \latestEdits{even when the parameter number} greatly exceeds the dataset size \citep{zhang2016understanding}.
DNNs are typically trained using Stochastic Gradient Descent (SGD),
in which model parameters at each optimisation step $\vw_{k+1}$ are updated using the gradient of the minibatch loss at the previous step $ L(\vw_{k})$:
\begin{equation}
	\vw_{k+1} = \vw_{k} - \alpha_{k}\nabla L(\vw_{k}),
\end{equation}
where $\alpha_{k}$ denotes the learning rate at iteration $k$. 
Whilst careful monitoring of the validation metrics, along with weight decay \citep{krogh1992simple}, layer-wise normalisation \citep{ioffe2015batch} and data-augmentation \citep{shorten2019survey,zhang2017mixup} help against over-fitting to the training data, the initial value of $\alpha_k$ and its schedule throughout training has a large generalisation impact \citep{jastrzkebski2017three,li2019towards}, making it a key hyperparameter.
Theoretical results for optimal asymptotic training set convergence prescribe a learning learning rate proportional to the inverse square root of the number of iterations \citep{nesterov2013introductory} or to decay at this rate \citep{duchi2018introductory}, however such schedules result in poor test set performance for DNNs. 
Furthermore, these convergence results only hold for the \textit{Iterate Average}, the arithmetic mean of all the iterates in the optimisation trajectory \citep{polyak1992acceleration}.  Curiously \citet{merity2017regularizing} and \citet{izmailov2018averaging} demonstrate that combining iterate averaging with large learning rate SGD increases DNN generalisation at the expense of training accuracy. However, quite why and how this works is still something of a mystery, limiting its widespread adoption.



One proposal in the literature to limit sensitivity to the learning rate and its schedule has been the development of \textit{adaptive gradient optimisers}, which invoke a per-parameter learning rate based on the history of gradients.
Popular examples include Adam \citep{kingma2014adam}, AdaDelta \citep{zeiler2012adadelta} and RMSprop \citep{tieleman2012lecture}. Ignoring momentum and explicit regularisation, the $k^{\mathrm{th}}$ iteration of a general adaptive optimiser is given by:
\begin{equation}
	\vw_{k+1} \leftarrow \vw_{k} - \alpha_k \mB^{-1} \nabla L_k(\vw_k),
\end{equation}
where the preconditioning matrix $\mB$ typically approximates curvature information.
Crucially, however, the generalisation of solutions found using adaptive methods, as measured in terms of test and validation error, significantly underperform SGD \citep{wilson2017marginal}. Due to this, state-of-the-art models, especially for image classification datasets such as CIFAR \citep{yun2019cutmix} and ImageNet \citep{xie2019selftraining,cubuk2019randaugment} are still trained SGD with momentum \citep{nesterov2013introductory}. Furthermore, despite IA being mentioned as a potential amendment in the original Adam paper \citep{kingma2014adam} and being required in the convergence proof \citep{reddi2019convergence}, it is not \latestEdits{widely}
used for computer vision problems.

\section{Contributions}
The key contribution of this paper are:
\begin{itemize}[leftmargin=0.1in, topsep=0.05pt, itemsep=0.01pt]
    \item We investigate the impact of IA on generalisation by considering extremely high-dimensional SGD on the quadratic model of the true risk perturbed by i.i.d. Gaussian noise, showing that the iterate average attains the global minimum, whereas the final point, despite multiple learning rate drops, or increases in batch size during training, does not.
    \item We extend the framework to a Gaussian Process perturbation model between the true and batch gradients. We find that as long as certain technical conditions (well met in practice) are satisfied, the simplified result holds. Crucially, distance in weight space or relative weight space (depending on kernel choice) are pivotal to the effect, \textit{justifying the need for large learning rates in practice}.
    \item We show that adaptive gradient methods have identical properties for the iterate average but we expect them to converge faster than their non adaptive counterparts.
    \item Motivated by these results, we consider the question \textit{why are adaptive methods not typically used in conjunction with iterate averaging}? We find that ineffective regularisation, which limits the effective distance and prediction diversity between the iterates, is the culprit and propose a simple and highly effective fix. We propose two Adam-based algorithms: \textbf{Gadam} and \textbf{GadamX}. They strongly outperform all baselines on all networks and datasets by large margins. GadamX achieves a Top-$1$ error of $22.69\%$ on ImageNet using ResNet-$50$, outperforming a well-tuned SGD baseline of $23.85\%$~\cite{pytorchImagenet}. To put this into perspective, \latestEdits{the gain attributed to widely adopted} cosine schedules only increases accuracy by $0.3\%$ \citep{bello2021revisiting}.
\end{itemize}
\emph{To the best of our knowledge, no existing work has shown that it is possible to outperform SGD (or indeed the stronger baseline of SGD with IA) on large scale problems using adaptive methods, so these results are, ipso facto, novel}.

\paragraph{Related Work \& Motivation:}
To the best of our knowledge there has been no explicit theoretical work analysing the generalisation benefit of Iterate Averaging. Whilst \citet{izmailov2018averaging} claim that Iterate Averaging leads to ``flatter minima which generalise better'', flatness metrics are known \latestEdits{have significant limitations as a proxy for generalisation}~\citep{dinh2017sharp}. 
Furthermore, we explicitly show in the \latestEdits{supplementary} 
that adaptive methods can find very sharp, highly generalising minima throwing the applicability of the analysis into question. \citet{martens2014new} show that the IA convergence rate for both SGD and second order methods are identical, but argue that second order methods have an optimal preasymptotic convergence rate on the convex quadratic. The analysis does not extend to \textit{generalisation} and no connection is made to adaptive gradient methods or the importance of the dimensionality of the problem, two major contributions of our work. Amendments to improve the generalisation of adaptive methods include switching between Adam and SGD \citep{keskar2017improving} and decoupled weight decay \citep{loshchilov2018decoupled}, limiting the extent of adaptivity \citep{chen2018closing,zhuang2020adabelief}. We incorporate these insights into our algorithms but significantly outperform them experimentally. The closest algorithmic contribution to our work is Lookahead \citep{zhang2019lookahead}, which combines adaptive methods with an exponentially moving average. We analyse this algorithm both theoretically (see supplementary) and practically and find significantly reduced theoretical benefit and weaker experimental performance. 

\section{Iterate Averaging: A New Theory for Generalisation}
The Iterate Average (IA) \citep{polyak1992acceleration} is the arithmetic mean of the model parameters over the optimisation trajectory $\vw_{\mathrm{avg}} = \frac{1}{n}\sum_{i}^{n}\vw_{i}$.
It is a classical variance reducing technique in optimisation with optimal asymptotic convergence rates and greater robustness to the choice of learning rate \citep{kushner2003stochastic}.  Indeed, popular regret bounds that form the basis of gradient-based convergence proofs \citep{duchi2011adaptive,reddi2019convergence} only imply convergence for the iterate average~\citep{duchi2018introductory}. 
For networks with batch normalisation~\citep{ioffe2015batch}, a naive application of IA (in which we simply average the batch normalisation statistics) is known to lead to poor results \citep{defazio2019ineffectiveness}.  However, by computing the batch normalisation statistics for the iterate average using a forward pass of the data at the IA point, \citet{izmailov2018averaging} show that the performance of small-scale image experiments such as CIFAR-10/100 and pretrained ImageNet can be significantly improved. Even for small experiments this computation is expensive, so they further approximate IA by taking the average at the end of each epoch instead of each iteration, which they call \emph{Stochastic Weight Averaging} (SWA). We show in the supplementary that experimentally the two approaches produce almost identical results, with SWA actually slightly outperforming IA. Since SWA can be seen as IA with a lower averaging frequency, we keep the terminology IA, however in our theoretical analysis we show explicitly include analysis for reduced frequency iterate averaging.

\subsection{A High Dimensional Geometry Perspective}
\label{subsec:theory}
We examine the variance reducing effect of IA in the context of a quadratic model for the true loss combined with additive perturbation models for the batch training loss. The theory we present is specifically high-dimensional (i.e. large \latestEdits{number of parameters,} $P$) and small batch size (small $B$) regime, which we term the ``deep learning limit''.
Intuitively, any given example from the training set $j \in \mathcal{D}$, will contain \textit{general features}, which hold over the data generating distribution and \textit{instance specific features} (which are relevant only to the training sample in question). For example, for a training image of a dog, we may have that: 
\begin{equation}
	\label{eq:instancespecific}
\overbrace{\underbrace{\nabla L_{sample}(\vw)}_\textrm{training set example}}^\textrm{dog $j$}  =\overbrace{ \underbrace{\nabla L_{true}(\vw)}_\textrm{general features}}^\textrm{$4$ legs, snout} + \overbrace{\underbrace{\vepsilon(\vw).}_\textrm{instance-specific features}}^\textrm{black pixel in top corner, green grass}
\end{equation}
Under a quadratic model of the \emph{true loss}\footnote{The loss under the expectation of the data generating distribution, rather than the loss over the dataset $L_{emp}(\vw_{k})$.} $L_{true}(\vw)=\vw^T\mH\vw$, where $\mH = \nabla^{2} L$ is the Hessian of the true loss w.r.t weights and we sample a mini-batch gradient of size $B$ at point $\vw \in \mathbb{R}^{P\times 1}$. The observed gradient is perturbed by an amount $\vepsilon(\vw)$ from the true loss gradient (due to instance specific features).  Under this model the component of the $\vw_{t}$'th iterate along the $j$'th eigenvector $\vphi_{j}$ of the true loss when running SGD with learning rate $\alpha$ can be written:
\begin{equation}
	\vw_{t}^{T}\vphi_{j} = (1-\alpha\lambda_{j})^{t}\vw_{0}^{T}\vphi_{j} - \alpha(1-\alpha\lambda_{j})^{t-1}\vepsilon(\vw_{1})^{T}\vphi_{j}...
\end{equation}
\latestEdits{Here} the $\lambda_j$ are the eigenvalues of $\mH$. \nickins{The simplest tractable model for the gradient noise $\vepsilon(\vw_t)$ is to assume they are i.i.d. isotropic multivariate Gaussians. In particular, this assumption removes any dependence on $\vw_t$ and and precludes the existence of any distinguished directions in the gradient noise. With this assumption, we can obtain Theorem \ref{theorem:shell}.}

\begin{theorem}
	\label{theorem:shell}
	Under the aforementioned assumptions, 
	\tiny
	\begin{equation}
	    \hspace{-0.4cm}
		\begin{aligned}
			& \mathbb{P}\bigg\{||\vw_{n}||-\sqrt{\sum_{i}^{P}w_{0,i}^{2}\mathrm{e}^{-2n\alpha\lambda_{i}}+P\frac{\alpha\sigma^{2}}{B}\bigg\langle\frac{1}{ \lambda(2-\alpha\lambda) }\bigg\rangle} \geq t \bigg\} \leq \nu ,
			\mathbb{P}\bigg\{||\vw_{\mathrm{avg}}||-\sqrt{\sum_{i}^{P}\frac{w_{0,i}^{2}}{\lambda_{i}^{2}n^{2}\alpha^{2}}+\frac{P\sigma^{2}}{Bn}\bigg\langle\frac{1}{ \lambda }\bigg\rangle}\geq t \bigg\} \leq \nu
		\end{aligned}
	\normalsize
	\end{equation}
	\normalsize
	where $\nu = 2\exp(-ct^{2})$, and $\langle \lambda^{k} \rangle = \frac{1}{P}\mathrm{Tr}\mH^{k}$. $B$ is the batch size (proof in supplementary).
\end{theorem}
The final iterate attains exponential convergence in the mean of $\vw_{n}$, but does not control the variance term. Whereas for $\vw_{\mathrm{avg}}$, although the convergence in the mean is worse (linear), the variance vanishes asymptotically -- this motivates \textit{tail averaging}, to get the best of both worlds. Another key implication of Theorem \ref{theorem:shell} lies in its dimensionality dependence on $P$. With $P$ being a rough gauge of the model complexity, this implies that \emph{in more complex, over-parameterised models, we expect the benefit of IA over the final iterate to be larger due to the corresponding variance reduction}. Note the limited extra improvement possible by simply increasing the batch size, compared to IA asymptotically.

\subsection{A dependent model for the perturbation:} We proceed now to propose a relaxation of the gradient perturbation independence assumption. Integrating (\ref{eq:instancespecific}) and writing the scalar perturbation term as $\eta$ (so $\nabla\eta = \vepsilon$), we model $\eta$ as a Gaussian process $\mathcal{GP}(m, k)$, where $k$ is some kernel function $\mathbb{R}^P\times \mathbb{R}^P\rightarrow\mathbb{R}$ and $m$ is some mean function\footnote{It is natural to take $m=0$ in a model for the sample perturbation, however retaining fully general $m$ does not affect our arguments.} $\mathbb{R}^P\rightarrow\mathbb{R}$. As an example, taking $k(\vw, \vw') \propto (\vw^T \vw')^p$ and restricting $\vw$ to a hypersphere results in $\vepsilon$ taking the exact form of a spherical $p$-spin glass, studied previously for DNNs \citep{choromanska2015loss,gardner1988optimal,mezard1987spin,ros2019complex,mannelli2019passed,baskerville2020loss, baskerville2021spin}. \emph{We are not} proposing to model the loss surface (sample or true) as a spin glass (or more generally, a Gaussian process), rather we are modelling the perturbation between the loss surfaces in this way.
\latestEdits{We emphasise that}
\emph{this model is a strict generalisation of the i.i.d. assumption above, and presents a rich but tractable model of isotropic Gaussian gradient perturbation without independence}.
\vspace{-5pt}
\paragraph{Notation:} With some variable $n\rightarrow\infty$, $f(n)\sim g(n)$ denotes asymptotic equivalence between $f$ and $g$, i.e. $(f(n) - g(n))/g(n)\rightarrow 0$, $f(n) = o(g(n))$ is shorthand for $f(n)/g(n)\rightarrow 0$ and $f(n) = \mathcal{O}(g(n))$ is shorthand for $f(n)/g(n)\rightarrow c$, for some constant $c>0$. In particular $\mathcal{O}(1)$ can be read as shorthand for any fixed non-zero constant, and $o(1)$ for any term which decays to $0$.

Following from our Gaussian process definition, the covariance of gradient perturbation can be computed using a well-known result (see \citet{adler2009random} equation 5.5.4), e.g.
 \begin{align} \label{eq:basic_gp_covar}
    Cov(\epsilon_i(\vw), \epsilon_j(\vw') ) = \partial_{w_i}\partial_{w'_j} k(\vw, \vw')
\end{align}
and further, assuming a stationary kernel $k(\vw, \vw') = k\left(-\frac{1}{2}||\vw - \vw'||_2^2\right)$ 
\begin{align}\label{eq:grad_gp_covar}
    & Cov(\epsilon_i(\vw), \epsilon_j(\vw') ) = (w_i - w'_i)(w'_j - w_j) k''\left(-\frac{1}{2}||\vw - \vw'||_2^2\right)  + \delta_{ij}k'\left(-\frac{1}{2}||\vw - \vw'||_2^2\right).
\nonumber
\end{align}
Thus we have a non-trivial covariance between gradient perturbation at different points in weight-space. This covariance structure can be used to prove the following variance reduction result.

\begin{figure}[t!]
	\begin{subfigure}[b]{0.24\linewidth}
		\includegraphics[width=\textwidth]{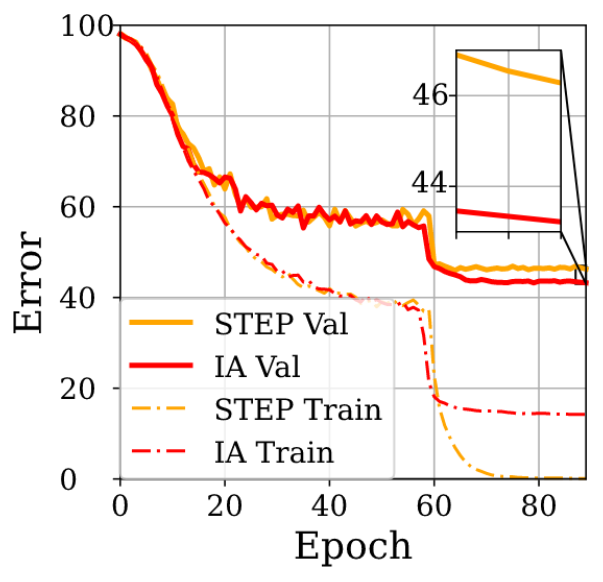}
		\caption{IA vs STEP schedule}
		\label{subfig:iavsstep}
	\end{subfigure}
	\begin{subfigure}[b]{0.24\linewidth}
		\includegraphics[width=\textwidth]{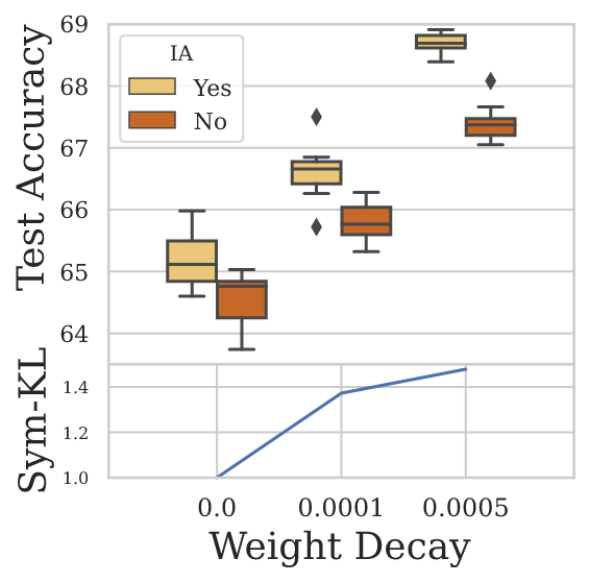}
		\caption{Weight Decay Impact}
		\label{subfig:wdswa}
	\end{subfigure}
	\begin{subfigure}[b]{0.24\linewidth}
		\includegraphics[width=\textwidth]{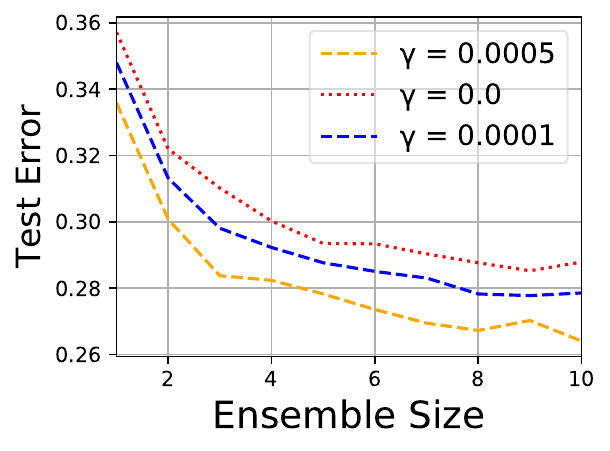}
		\vspace{-10pt}
		\caption{Network Ensemble}
		\label{subfig:ensemblevgg16}
	\end{subfigure}
	\begin{subfigure}[b]{0.24\linewidth}
		\includegraphics[width=\textwidth]{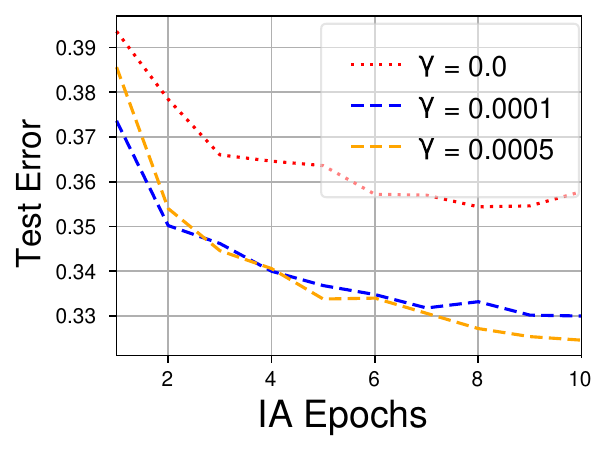}
		\vspace{-10pt}
		\caption{Iterate Averaging}
	    \label{subfig:swavgg16}
	\end{subfigure}
	\caption{(a) \latestEdits{STEP (learning rate decay)} and IA Train/Val error. Both approaches reduce Val error, but IA by a greater margin. (b) Effect of weight decay on held out test error for IA/sharp learning rate decay solutions. Greater weight decay increases the margin of IA improvement. The lower subplot shows the average symmetric KL-divergence between IA solutions. (c-d) Test error improvement with differing degrees of regularisation $\gamma$ for (c) network ensembling and (d) IA.}
	\vspace{-10pt}
\end{figure}

\begin{theorem}\label{theorem:dependent_noise_shell}
Let $\vw_n$ and $\vw_{avg}$ be defined as in Theorem \ref{theorem:shell} and let the gradient perturbation be given by the covariance structure in (\ref{eq:grad_gp_covar}). Assume that the kernel function $k$ is such that $k'(-x^2)$ and $x^2k''(-x^2)$ decay as $x\rightarrow\infty$, and define $\sigma^2B^{-1} = k'(0)$. Assume further that $P\gg \log n$. Let $\delta=o(P^{1/2})$. Then $\vw_n$ and $\vw_{avg}$ are multivariate Gaussian random variables and, with probability which approaches unity as $P, n\rightarrow\infty$ the iterates $\vw_t$ are all mutually at least $\delta$ apart and 

\small
\begin{alignat}{2}
    &\mathbb{E}w_{n,i} \sim e^{-\alpha\lambda_i n}w_{0,i} ,
    &&\frac{1}{P}\Tr Cov(\vw_n) \sim \frac{\alpha\sigma^2}{B}\left\langle\frac{1}{\lambda(2-\alpha\lambda)}\right\rangle,\\
    &\mathbb{E}w_{avg, i} \sim \frac{1-\alpha\lambda_i}{\alpha\lambda_i n} w_{0,i},
    &&\frac{1}{P}\Tr Cov(\vw_{avg}) \leq \frac{\sigma^{2}}{Bn}\left\langle\frac{1}{ \lambda }\right\rangle +  \mathcal{O}(1)\Bigg(k'(-\frac{\delta^2}{2}) + P^{-1}\delta^2k''(-\frac{\delta^2}{2})\Bigg)\label{eq:ia_gp_var}.
\end{alignat}
\end{theorem}
\latestEdits{Note} that Theorem \ref{theorem:dependent_noise_shell} is a generalisation of Theorem \ref{theorem:shell} to the context of our dependent perturbation model. 
The proof of the theorem (in the supplementary) makes plain the importance of $P\rightarrow\infty$ for this result, i.e. our result is specifically applicable to iterate averaging in the \emph{Deep Learning Limit} of very many model parameters. Indeed, the IA variance term (\ref{eq:ia_gp_var}) cannot be said to decay as $n^{-1}$ (as in Theorem \ref{theorem:shell}) as there are the additional terms arising from gradient perturbation correlations which decrease only with the distance between the $\vw_{i}$ and can be shown to decay only in the $P\rightarrow\infty$ limit. \textit{This establishes the importance of large learning rates in conjunction with Iterate Averaging, as the iterates must be sufficiently ``far apart" to retain the independence benefit and variance reduction. Furthermore, the importance of $P \gg \log n$ indicates that perhaps averaging less frequently than every iteration could be beneficial to generalisation, something we observe in practice (see supplementary)}.
\nickins{Explicitly, suppose that we average only every $\kappa$ iterations (where $\kappa$ is fixed). Intuitively, the first term in the covariance  in (\ref{eq:ia_gp_var}) is an ``independence term'', i.e. it is common between Theorems \ref{theorem:shell} and \ref{theorem:dependent_noise_shell} and represents the simple variance reducing effect of averaging. The second variance term in (\ref{eq:ia_gp_var}) comes from dependence between the iterate gradient perturbations. In the supplementary material
we establish the that first term increases roughly by a factor $\kappa$ due to striding, whereas the second term is unaffected (to leading order). We see therefore that an independent model for gradient perturbation would predict an unambiguous inflationary effect of strided IA on variance. However introducing dependence in the manner that we have predicts a more nuanced picture, where increased distance between weight iterates can counteract the simple ``independent term'' inflationary effect of striding, leaving open the possibility for striding to improve on standard IA for the purposes of generalisation, as we observe experimentally in the Supplementary.}
Finally, note that $P\gg \log{n}$ is a perfectly reasonable condition in the context of deep learning. E.g. for a ResNet-50 with $P\approx 25\times 10^{6}$, violation of this condition would require $n > 10^{10^{7}}$. A typical ResNet schedule on ImageNet has $< 10^{6}$ total steps. 
\vspace{-5pt}
\paragraph{Validation of Theory:}
To better understand the effect of the large learning learning rate on generalisation, we train a VGG-$16$ network~\citep{simonyan2014very} with no data augmentation/batch normalisation (to isolate the overfitting effect from reducing the learning rate) with a learning rate of $\alpha=0.05$. Replacing the learning rate drop (performed at epoch $60$ by a factor of $10$ with weight decay $\gamma = 0.0005$) with IA at the same point, we find that the test error is reduced by a greater margin ($\approx 2\%$), shown in Figure~\ref{subfig:iavsstep}. We note that IA improves over the SGD learning rate equivalent for all values of weight decay, with results for $10$ seeds shown in Fig~\ref{subfig:wdswa} and hence this argument is independent or explicit regularisation as indicated by Theorem~\ref{theorem:dependent_noise_shell}. For our Deep Neural Network experiments, we find that the best IA optimiser improvement over its base optimiser is proportional to the number of parameters $P$ as shown in Fig~\ref{subfig:improvementp} and predicted by Theorem~\ref{theorem:shell}. This theorem and experimental validation thereof translates into the following advice for practitioners: \textit{In the Deep Learning limit (large $P$ and small relative $B$) one should keep the learning rate high and use iterate averaging instead of sharply dropping the learning rate!}

\subsection{A Closer Look at IA and the Importance of Regularisation}
\label{subsec:wdia}
\citet{izmailov2018averaging} argue under a linearisation assumption that IA can be seen as approximate model ensembling. Since averaging only improves test performance for sufficiently uncorrelated models (through a reduction in variance of the ensemble), we must ensure sufficiently diverse models at each epoch through our training procedure. 
We note from Fig~\ref{subfig:swavgg16}, that unlike model ensembling (shown in Fig~\ref{subfig:ensemblevgg16}), the IA improvement is strongly dependent on the use of weight decay. 
We show the difference between IA and sharply decaying learning rate schedules (which mirror conventional setups) over $10$ seeds as a function of weight decay coefficient in Fig~\ref{subfig:wdswa}. The margin of improvement from IA over sharply decaying schedules
steadily increases with regularisation with $\gamma=0.0005$ delivering a greater final validation accuracy at the final IA point despite starting from a lower starting point than $\gamma=0.0001$. 
To explicitly show that weight decay encourages greater diversity in the iterates we calculate the symmetrised KL-divergence $\frac{1}{2}(\sum p(x)\log\frac{p(x)}{q(x)}+q(x)\log\frac{q(x)}{p(x)})$ over the entire test set between the softmax outputs of the IA iterates, taking an average for each weight decay value, shown (normalising the $0$ weight decay value to $1$) in the lower subfigure in Fig~\ref{subfig:wdswa}. As expected greater weight decay gives greater solution diversity.
\paragraph{Distance in weight space or relative distance?} It is natural to ask given the results from \ref{subsec:theory} and Theorem~\ref{theorem:dependent_noise_shell}, which require distance in weight-space between iterates to achieve the variance reduction, whether this is the reason for the efficacy of larger weight decay? For learning rate $\alpha$ and weight decay $\gamma$ we move a distance $\alpha \nabla L(\vw) - \alpha \gamma \vw$ in weight space. Intuitively, since random vectors in high dimensions are nearly orthogonal with high probability \citep{vershynin2018high}, we expect the distance in weight space to move a distance $\alpha\sqrt{(\nabla L(\vw))^{2}+\gamma^{2}\vw^{2}}$, which is larger for $\gamma>0$. Conversely, we expect $||\vw||^{2}$ to be smaller for smaller $\gamma$ and the gradients also to be smaller \citep{granziol2020flatness}. For the experiment (with equal learning rates) shown in Fig~\ref{subfig:wdswa}, the average distances between the IA epochs for weight decay values $\gamma = \{0,0.0001,0.0005\}$ are $17.7,14.9,13.9$ respectively. We note however that the relative distance, when normalised by the average weight norm increases successively $0.11,0.13,0.22$. This begs the question whether Theorem~\ref{theorem:dependent_noise_shell} can be extended to include a notion of relative distance.
\begin{theorem}\label{theorem:dependent_noise_shell_relative}
Let $\vw_n$ and $\vw_{avg}$ be defined as in Theorem \ref{theorem:dependent_noise_shell} and let the gradient noise be given by the basic covariance structure in (\ref{eq:basic_gp_covar}).
Let the kernel function be of the form \begin{align*}
    k(\vw, \vw') = k\left(-\frac{||\vw - \vw'||_2^2}{||\vw||_2^2}\right)
\end{align*}and assume that the kernel function $k$ is such that $k'(-x^2)$ and $x^2k''(-x^2)$ decay as $x\rightarrow\infty$, and define $\sigma^2B^{-1} = k'(0)$. Assume further that $P\gg \log n$. Then the result of Theorem \ref{theorem:dependent_noise_shell} holds.
\end{theorem}
In Theorem \ref{theorem:dependent_noise_shell} it is the absolute distance between iterates that determines the strength of dependence. In Theorem \ref{theorem:dependent_noise_shell_relative}, the gradient noise covariance is sensitive instead to the the \emph{relative} distance. This notion can be easily extended to products of weights in different layers, which is known to be invariant with the use of batch normalisation \citep{ioffe2015batch}, used in many modern networks.

\section{Adaptive Gradient Methods with Iterate Averaging}

\begin{figure}[h!]
	\hspace{-10pt}
	\begin{subfigure}[b]{0.3\linewidth}
		\includegraphics[width=\linewidth]{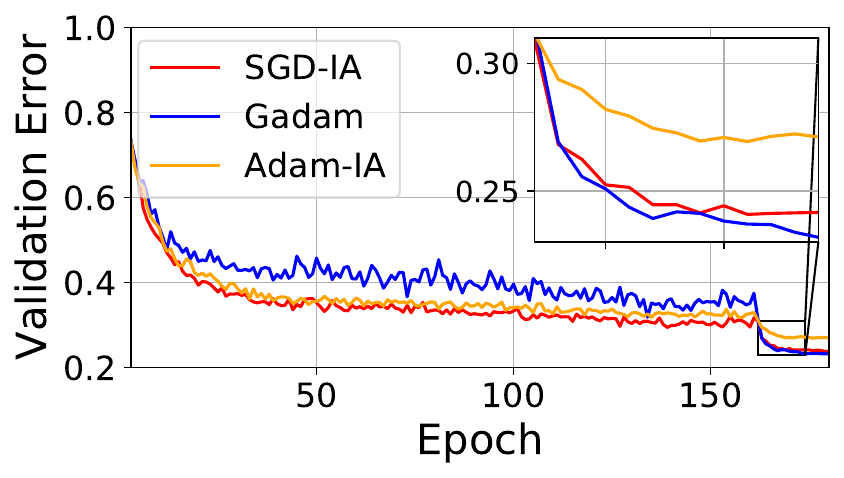}
		\caption{P$110$ Val. Error}
		\label{subfig:adamia}
	\end{subfigure}
	\begin{subfigure}[b]{0.25\linewidth}
	\begin{tiny}
        \begin{tabular}{l@{\hspace{0.8\tabcolsep}}c@{\hspace{0.8\tabcolsep}}c@{\hspace{0.8\tabcolsep}}c@{}}
        \toprule
        Optim & SGD-IA & Adam-IA & Gadam \\
        \midrule
        $\mathcal{D}(p||q)$ & 20.2 & 19.1 & 22.8 \\
        $||p-q||^{2}$ & 18.2 & 18.0 & 18.6  \\
        $\Delta E$ & 9.0 & 5.8 & 11.9 \\
        \bottomrule
        \multicolumn{3}{l}{Values except $\Delta E$ in units of $1000$.} \\
        \end{tabular}
        \vspace{10pt}
    \end{tiny}
    \caption{Solution diversity}
    \end{subfigure}
    \hspace{15pt}
    	\begin{subfigure}[b]{0.22\linewidth}
		\includegraphics[width=\textwidth]{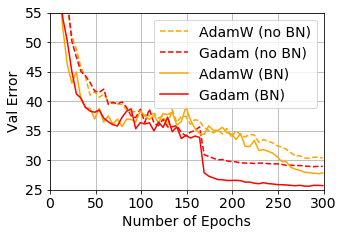}
		\caption{Val Error}
		\label{subfig:bnvsnobn}
	\end{subfigure}
	\hspace{-5pt}
	\begin{subfigure}[b]{0.19\linewidth}
		\includegraphics[trim={1cm 0 0 0},clip,width=\textwidth]{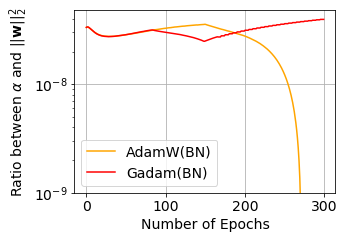}
		\caption{${\alpha}/{||\vw||^2}$}
		\label{subfig:effectivealpha}
	\end{subfigure}
\caption{(a) Validation error for the PreResNet-$110$ on CIFAR-$100$ for various optimisers using IA and (b) the solution diversity given as the symmetrised KL $\mathcal{D}$ or total variation distance calculated on the test set and the change in validation error $\Delta E$ for the final IA point.  Val. error and effective learning rate $\frac{\alpha}{||\vw||^2}$ of VGG-16 on CIFAR-100 with and without BN.}
\label{tab:metrics}    
\vspace{-5pt}
\end{figure}
Naively combining IA with Adam is ineffectual, as shown in Fig.~\ref{subfig:adamia}. Despite the same $L_2$ regularisation ($0.0001$), the error drop is significantly less than for SGD-IA. Following our intuition from Sec~\ref{subsec:wdia}, we consider whether the problem could be that overly correlated solutions form the IA due to ineffective regularisation. As shown in Tab.~(b) of Fig. \ref{tab:metrics}, both the symmetrised KL divergence $\mathcal{D}(p||q)$ and total variation distance $||p-q||^{2}$ (calculated between all epochs using IA at the end of training and then averaged) are lower for Adam-IA than for SGD-IA. For adaptive optimisers, $L_2$ regularisation is not equivalent to weight decay \citep{zhang2018three,loshchilov2018decoupled}, with weight decay generalising better, known as \emph{AdamW}. For AdamW with decoupled weight decay $0.25$ the solution diversity increases to even beyond that of SGD-IA. It is accompanied by a greater drop in validation error, even outperforming SGD-IA. We term this combination of AdamW + IA \emph{Gadam} to denote a variant of Adam that generalises. Previous work has shown that limiting the belief in the curvature matrix of Adam improves generalisation \citep{zhuang2020adabelief,chen2018closing}, hence we also incorporate such a partially adaptive Adam into our framework and term the resulting Algorithm \emph{GadamX}.
For convolutional neural networks using batch normalisation, the effective learning rate is proportional to $\alpha_{\mathrm{eff}} \propto \frac{\alpha}{||\vw||^2}$ \citep{hoffer2018norm}. With batch normalisation, the output is invariant to the channel weight norm, hence the weight changes only with respect to the direction of the vector. Since the effective weight decay depends on the (effective) learning rate, we expect this to lead to more regularised solutions and better validation error.
To test this hypothesis, we train a VGG-16 network on CIFAR-100 with and without BN (see Figs \ref{subfig:bnvsnobn},\ref{subfig:effectivealpha}): for an identical setup, 
the margin of improvement of Gadam over AdamW is much larger with BN. We note that \latestEdits{while} Gadam keeps the effective learning rate $\frac{\alpha}{||\vw||^2}$ high, in scheduled AdamW it quickly vanishes once we start learning rate decay. We were unable to compensate with learning rate scheduling, underscoring the importance of appropriate weight decay.

\paragraph{Extension of theoretical framework to weight decay and adaptive methods:} 
Introducing decoupled weight decay (strength $\gamma$) and moving the true loss minimum away from the origin $L_{\mathrm{true}}(\vw) = (\vw-\vw^*)^T\mH(\vw-\vw^*)$. The update rule for adaptive methods is then
\begin{align}\label{eq:adam_update}
    \vw_t = \left(1-\alpha\gamma - \alpha \tilde{\mH}_t^{-1}\mH\right) \vw_{t-1} + \alpha \mH \vw^* - \alpha \vepsilon(\vw_{t-1})
\end{align}
where $\tilde{\mH}_t^{-1}$ is the approximation to the true loss Hessian used at iteration $t$, and is taken to be diagonal in the eigenbasis of $H$, with eigenvalues $\tilde{\lambda}_i^{(t)}+\varepsilon$, where $\varepsilon$ is the standard tolerance parameter. One could try to construct the $\tilde{H}_t^{-1}$ from the Gaussian process loss model, so making them stochastic and covariate with the gradient noise, however we do not believe this is tractable. Instead, let us heuristically assume that, with high probability, $\tilde{\lambda}_i^{(t)}$ is close to $\lambda_i$ for large enough $t$ and all $i$. If we take a large enough $\zeta$ this is true even for SGD and we expect Adam to better approximate the local curvature matrix than SGD \citep{granziol2020explaining}. This results in the following theorem.

\begin{theorem}\label{theorem:shell_gadam}
Fix some $\zeta > 0$ and assume that $|\tilde{\lambda}_i^{(t)} - \lambda_i| < \zeta$ for all $t \geq n_0$, for some fixed $n_0(\zeta)$, with high probability. Use the update rule (\ref{eq:adam_update}). Assume that the $\lambda_i$ are bounded away from zero and $\min_i\lambda_i > \zeta$. Further assume $c(\gamma + \varepsilon + \zeta) < 1$, where $c$ is a constant independent of $\varepsilon, \zeta, \gamma$ and is defined in the proof. Let everything else be as in Theorem \ref{theorem:dependent_noise_shell}. Then there exist constants $c_1, c_2, c_3, c_4>0$ such that, with high probability,
\small
\begin{align}
    &|\mathbb{E}w_{n,i}-w^*_i| \leq e^{-\alpha(1 +\gamma -c(\varepsilon+\zeta))n}w_{0,i} + c_1(\varepsilon +\zeta + \gamma)\label{eq:thm4_1} \\
    &\left|\frac{1}{P}\Tr Cov(\vw_n)- \frac{\alpha\sigma^2}{B(2-\alpha)}\right| \leq   c_2(\varepsilon + \zeta + \gamma) + o(1),\label{eq:thm4_2}\\
    &|\mathbb{E}w_{avg, i} - w^*_i| \leq \frac{1-\alpha(1 +\gamma - c(\varepsilon + \zeta ))}{\alpha(1 + \gamma - c(\varepsilon+\zeta)) n} (1 + o(1))w_{0,i} + c_3(\varepsilon + \zeta + \gamma)\label{eq:thm4_3}\\
    &\left|\frac{1}{P}\Tr Cov(\vw_{avg}) - \frac{\sigma^{2}}{Bn} -  \mathcal{O}(1)\Bigg(k'(-\frac{\delta^2}{2}) + P^{-1}\delta^2k''(-\frac{\delta^2}{2})\Bigg)\right| \leq c_4 (\gamma, + \zeta + \epsilon).\label{eq:thm4_4}
\end{align}
\end{theorem}
Theorem \ref{theorem:shell_gadam} demonstrates the same IA variance reduction as seen previously but in this more general context of weight decay and adaptive optimisation. As expected, improved estimation of the true Hessian eigenvalues (i.e. smaller $\zeta$) reduces the error in recovery of $\vw^*$. Moreover, increasing the weight decay strength $\gamma$ decreases the leading order error bounds in (\ref{eq:thm4_1}) and (\ref{eq:thm4_3}), but only up to a point, as the other error terms are valid and small only if $\gamma$ is not too large.

\section{Experiments}
\label{sec:experiments}
\begin{figure*}[h!]
	\centering
	\begin{subfigure}[b]{0.24\textwidth}
		\includegraphics[width=\textwidth]{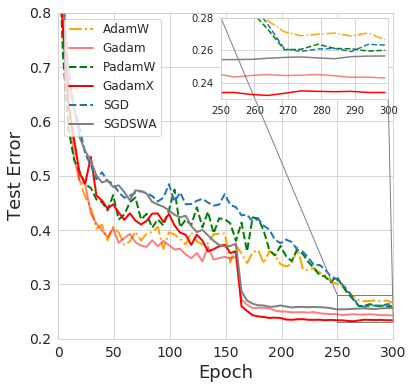}
		\caption{VGG-16}
	\end{subfigure}
	\begin{subfigure}[b]{0.24\textwidth}
		\includegraphics[width=\textwidth]{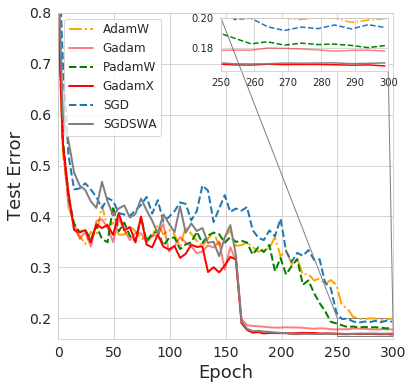}
		\caption{ResNeXt-29}
	\end{subfigure}
	\begin{subfigure}[b]{0.24\linewidth}
		\includegraphics[width=\textwidth]{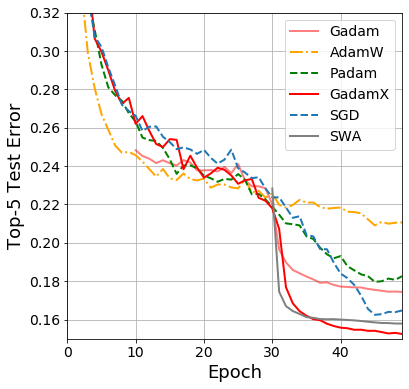}
		\caption{WRN-28-10}
		\label{subfig:imagenet}
	\end{subfigure}
	\begin{subfigure}[b]{0.22\textwidth}
		\includegraphics[width=\textwidth]{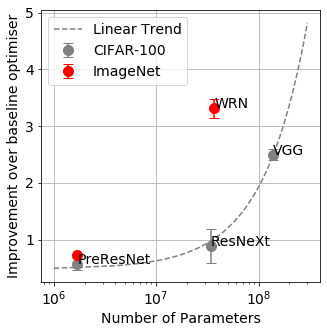}
		\caption{Improvement vs. $P$}
		\label{subfig:improvementp}
	\end{subfigure}
	\caption{(a-b) Top $1$ Test error on CIFAR-$100$, (c) Top-$5$ Test Error on ImageNet-$32$ and (d) IA test improvement over its base optimiser against number of parameters.}\label{fig:c100}
	\vspace{-10pt}
\end{figure*}

\subsection{Image Classification on CIFAR and Downsampled 32x32 ImageNet Datasets}
\label{subsec:cifarexp}
Here we consider VGG-16, Preactivated ResNet (PRN) and ResNeXt \citep{simonyan2014very, he2016identity, xie2017aggregated} on CIFAR datasets \citep{krizhevsky2009learning}, with CIFAR-100 results in Fig. \ref{fig:c100} along with CIFAR-10 results in Table \ref{tab:cifar10and100}.
As AdamW always outperforms Adam in our experiments, the curves for the latter are omitted in the main text; we detail these results in the supplementary. The results show that optimisers with IA (SWA, Gadam and GadamX) invariably improve over their counterparts without, and
GadamX always delivers the strongest performance. Without compromising convergence speed, Gadam outperforms tuned SGD and Padam - suggesting that solutions found by adaptive optimisers do not necessarily generalise more poorly, as suggested in the literature \citep{wilson2017marginal}. Indeed, any generalisation gap seems to be closed by the using IA and an appropriately implemented weight decay. We emphasise that results here are achieved \textbf{without} tuning the point at which we start averaging $T_{\mathrm{avg}}$; if we allow crude tuning of $T_{\mathrm{avg}}$, on CIFAR-100 GadamX achieves 77.22\% (VGG-16) and 79.41\% \footnote{As opposed to $77.90\%$ without tuning.}
(PRN-110) test accuracy respectively, which to our knowledge are the best reported performances on these architectures.
We show results on ImageNet 32$\times$32 \citep{chrabaszcz2017downsampled} in Fig. \ref{subfig:imagenet}. While Gadam does not outperform our strong SGD baseline, it nevertheless improves upon AdamW greatly and posts a performance stronger than the SGD baseline in literature with identical \citep{chrabaszcz2017downsampled} and improved \citep{mcdonnell2018training} setups. Finally, GadamX performs strongly, outperforming more than 3\% compared to the baseline \citep{chrabaszcz2017downsampled} in Top-5 accuracy. We run each experiment three times with mean and standard deviation reported. In this section, all non-IA baselines are tuned rigorously with proper schedules for fair comparisons\footnote{In image classification, we use the \emph{linear schedule}, which both performs better than usual step schedule (see supplementary) and is consistent with \citep{izmailov2018averaging}.}, and we also include the results reported in the previous works in 
the supplementary, where we also include all the implementation details. \emph{In summary, to showcase the out of the box efficacy of our optimisers, we do not experiment on learning rates that vary from the base optimiser for CIFAR and downsampled ImageNet experiments. For Gadam we use an initial learning rate of $0.001$ (a default used by AdamW) and for GadamX we use an initial learning rate of $0.1$ (default for Padam~\citep{chen2018closing}).}

\subsection{ImageNet Experiments}
\label{sec:fullimagenet}
We compare against step learning rate decay (factor of $10$ every $30$ epochs) and linear schedule for SGD and AdamW for $90$ epochs \citep{he2016deep}, with respective initial learning rates $\alpha=0.1,0.001$ and weight decays $10^{-4},10^{-2}$ on ImageNet~\citep{russakovsky2015imagenet}. We combine LookAhead \citep{zhang2019lookahead} with gradient centralisation \citep{yong2020gradient} as a high performance adaptive baseline ~\emph{Ranger} \citep{Ranger}, also using step decay. We search for the best performing initial learning rates for SWA, GadamX and Ranger by factors of $3$ i.e $0.001,0.003$ in either direction (increase/decrease) until we find a local maximum in performance, otherwise leaving settings as in \ref{subsec:cifarexp}. Following \citet{granziol2020explaining,choi2019empirical}, we experiment with setting the numerical stability coefficient to $\delta=10^{-4}$ instead of $10^{-8}$ for GadamX and attempt an SGD like procedure for Gadam where we train with $\alpha=0.5, \delta=1, \gamma=10^{-4}$. Due to poor \say{out of the box} performance of SWA, we repeat the logarithmic grid search procedure on the IA learning rate for SWA. 
We report results in Table \ref{tab:imagenetfull}, where we see Gadam(X) strongly out-performing all baselines. We do not include the ResNet-$18$ as AdamW outperforms SGD with $69.92\%$ top-$1$ accuracy over $69.72\%$, hence not a useful test-case for
analysing the \emph{adaptive generalisation gap}, prevalent in deeper models. Gadam nonetheless improves on this \latestEdits{attaining} $70.11\%$. Whilst we find that step/linear scheduling is less effective for AdamW/SGD, attaining $73.68/75.52 \%$ respectively on the ResNet-$50$. Since these are small difference we don't consider scheduling to be a major factor in our outstanding results. We report the (slightly less) strong results of Gadam(X), without epsilon tuning in the Supp matt. 
\begin{figure}[t!]
	\begin{subfigure}[b]{0.24\linewidth}
		\includegraphics[width=\textwidth]{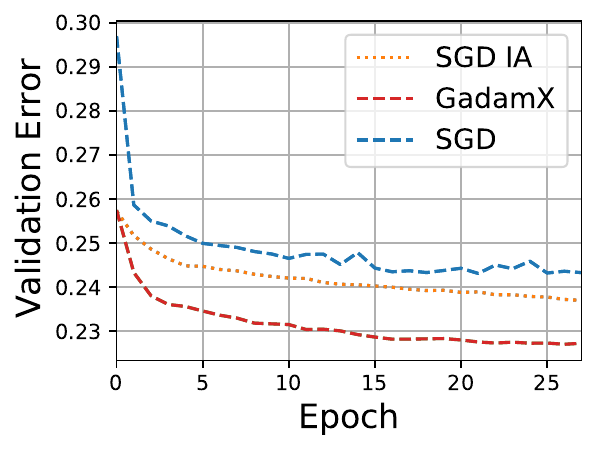}
		\caption{ResNet-$50$}
		\label{subfig:resnet50}
	\end{subfigure}
	\begin{subfigure}[b]{0.24\linewidth}
		\includegraphics[width=\textwidth]{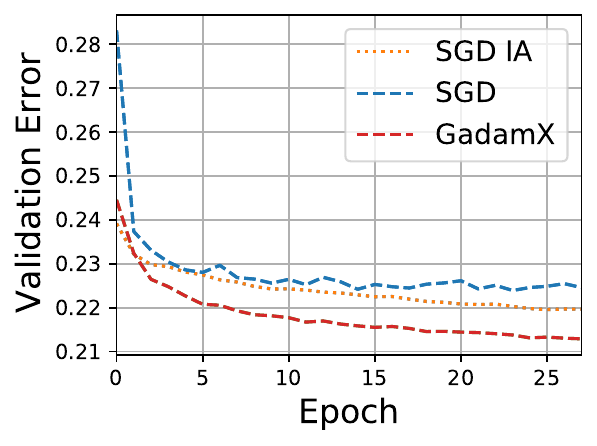}
		\caption{ResNet-$101$}
		\label{subfig:resnet101}
	\end{subfigure}
	\begin{subfigure}[b]{0.45\linewidth}
	    	\begin{scriptsize}
				\begin{tabular}{llcc}
					\toprule
					
					Data-set & optimiser &\multicolumn{2}{c}{Perplexity} \\
					& & Validation & Test \\
					\midrule
					PTB & ASGD & 64.88$\pm0.07$ & 61.98$\pm0.19$ \\
					& Adam & 65.96$\pm0.08$ & 63.16$\pm0.24$ \\
					& Gadam & \textbf{61.35}$\pm0.05$& \textbf{58.77}$\pm0.08$\\
					& GadamX & 63.49$\pm0.19$ & 60.45$\pm0.04$\\
					\bottomrule
				\end{tabular}
			\end{scriptsize}
	\caption{Validation and Test Perplexity on Word-level Language Modelling.}
	\label{tab:lstm}
	\end{subfigure}
	\caption{(a-b) Final ImageNet epochs, showing the improvement of both SGD with Iterate Averaging (SGD IA) and our proposed GadamX optimiser over the SGD step-schedule in Top-$1$ validation error. (c) LSTM Penn Treebank Experimental results.}
	\label{fig:iavsstep}    
	\vspace{-15pt}
\end{figure}
\begin{table}[!htb]
    \begin{minipage}{.5\linewidth}
      \caption{ImageNet Results}
      \label{tab:imagenetfull}
      \centering
\begin{tiny}
\begin{tabular}{llcc}
\toprule
Architecture & Optimiser & Top-1 & Top-5 \\
\midrule 
ResNet-50 & SGD(step) &75.63& 92.67  \\
& SWA & 76.32& 93.15 \\
& AdamW(lin) & 74.04 & 91.57 \\
& Ranger & 75.64 & 92.53 \\
& Gadam & 76.79 & 93.21\\
& GadamX & \textbf{77.31} & \textbf{93.47}\\
\midrule
ResNet-101 & SGD (step) & 77.37&93.78 \\
 & SWA & 78.08 &  93.92 \\
 & AdamW(lin) & 74.48 & 91.82 \\
& Ranger & 75.62 & 92.42 \\
 & Gadam & 78.53 & \textbf{94.29}\\
& GadamX & \textbf{78.72} & 94.18\\
\bottomrule
\end{tabular}
\end{tiny}
    \end{minipage}%
    \begin{minipage}{.5\linewidth}
      \centering
        \caption{CIFAR-$10$/$100$ Results}
        \label{tab:cifar10and100}
\begin{tiny}
\begin{tabular}{llccc}
\toprule
Architecture & Optimiser & C-$10$ Test Accuracy & C-$100$ Test Accuracy\\
\midrule
VGG-16    & SGD & 94.14$\pm0.37$ &  74.15$\pm 0.06$ \\
& SWA & 94.69$\pm0.36$ &  74.57$\pm 0.27$ \\
& Adam(W) & 93.90 $\pm0.11$ & 73.26$\pm 0.30$ \\
& Padam(W) & 94.13 $\pm0.06$ & 74.56$\pm 0.19$ \\
& Gadam & 94.62$\pm0.15$ & 75.73$\pm 0.29$ \\
& GadamX & \textbf{94.88}$\pm0.03$& \textbf{76.85}$\pm 0.08$\\
\midrule
PRN-110 & SGD & 95.40$\pm0.25$&77.22$\pm 0.05$ \\
& SWA & 95.55$\pm0.12$& \textbf{77.92}$\pm 0.36$ \\
& Adam(W) & 94.69$\pm0.14$ & 	75.47$\pm 0.21$\\
& Padam(W) & 95.28$\pm0.13$ & 77.30
$\pm 0.11$ \\
& Gadam & 95.27$\pm0.02$ & 77.37$\pm 0.09$ \\
& GadamX & \textbf{95.95$\pm0.06$} & 77.90 $\pm 0.21$ \\
\bottomrule
\end{tabular}
\end{tiny}
\end{minipage} 
\end{table}
\subsection{Beyond Computer Vision: PTB LSTM}
We run word-level language modelling using a 3-layer Long-short Term Memory (LSTM) model \citep{gers1999learning} on the PTB dataset \citep{marcus1993building} and the results are in Table \ref{tab:lstm}. Remarkably, Gadam achieves a test perplexity of 58.77 (58.61 if we tune $T_{\mathrm{avg}}$), better than the baseline NT-ASGD in \citet{merity2017regularizing} that \textit{runs an additional 300 epochs} on an identical network. 
\section{Conclusion}
We propose a Gaussian Process perturbation between the batch and true risk surfaces and derive the phenomenon of improved generalisation for large learning rates and larger weight decay when combined with iterate averaging observed in practice. We extend this formalism to include adaptive methods and show that we expect further improvement when using adaptive algorithms. Based on this theory we develop two adaptive algorithms,
Gadam and GadamX, variants of Adam with iterate averaging. We extensively validate Gadam and GadamX on computer vision tasks and a single natural language experiments, showing strong performance against baseline and state of the art. Another interesting consequence of our work is that in all our experiments \emph{the last iterate is the best.} Unlike SGD, where the epoch of best test/validation error is typically not the last and techniques such as early stopping are often employed, we find consistent near-monotonic improvements in test/validation error using our algorithms. We also find from preliminary analysis that our algorithms require less hyper-parameter tuning than SGD and variants thereof.
This may be of interest for practitioners that want to get good results fast, as opposed to state of the art slowly.
\paragraph{Limitations:}
Our theoretical framework encapsulates the dependence between risk surfaces as a Gaussian process; in practice higher order terms may be non-negligible. Whilst we report positive results in terms of less tuning and the last iteration giving the best validation error, further experimentation is required to fully establish these phenomena. Whilst we are confident due to our theoretical analysis/ experiments that adaptive methods with IA outperform non adaptive methods, it is possible that combining schedules on alternative networks and datasets could refute this claim. Notably certain heuristics such as ``super-fast convergent" schedules have not been trialled. We also do not address \emph{why} less aggressive curvature matrices benefit generalisation for computer vision.
\paragraph{Broader Impact:}
Generalisation is a key metric in machine learning applications, ranging from speech recognition to drug discovery~\citep{lecun2015deep}. We believe the broader impact of our work is enabling more efficient exploration and development of such applications, which can produce complex and potentially negative effects on society as a whole (e.g. the deployment of speech recognition for population surveillance). Beyond this, we hope that the robustness in terms of converging to high quality solutions even with less tuning of our proposed optimisation algorithms may enable researchers to reduce their energy consumption (and thus potentially their environmental impact) relative to alternatives such as hand-tuned SGD, which may require additional training runs to find good hyperparameter choices.




\bibliography{refs}
\bibliographystyle{icml2021}
\input{ICML-1394-submission/neurips_2021_checklist}

\appendix


\section{Gadam/GadamX Algorithm}
Here we present the full Gadam/GadamX algorithm. Note that for simplicity, in Algorthm \ref{alg:Gadam} we present a Polyak-style averaging of every iteration, in practice we find both practical and theoretical results why averaging \textit{less} frequently is almost equally good, if not better. We include a discussion on this in Appendix \ref{sec:avgfreq}.

\begin{algorithm}[H]
	\caption{Gadam/GadamX}
	\label{alg:Gadam}
	\begin{algorithmic}
		\REQUIRE initial weights $\theta_0$; learning rate scheduler $\alpha_t = \alpha(t)$; momentum parameters $\{\beta_{1}, \beta_2\}$ (Default to $\{0.9, 0.999\}$ respectively); 
		partially adaptive parameter $p \in [0, 0.5]$ Default to $\{0.125,0.5\}$ for \{GadamX, Gadam\}; decoupled weight decay $\gamma$; averaging starting point $T_{\mathrm{avg}}$; tolerance $\epsilon$ (default to $10^{-8}$)
		\ENSURE Optimised weights $\Tilde{\theta}$
		\STATE Set $\vm_0 = 0, \vv_0 =0, \hat{\vv_0} = 0, n_{\mathrm{models}} = 0$.
		\FOR{t = 1, ... T}
		\STATE $\alpha_t = \alpha(t)$
		\STATE $\vg_t = \nabla f_t(\theta_t)$
		\STATE $\vm_t = \beta_1\vm_{t-1} + (1 - \beta_{1})\vg_t / (1 - \beta_{1}^t)$
		\STATE $\vv_t = \beta_2\vv_{t-1} + (1 - \beta_2)\vg_t^2 / (1 - \beta_{2}^t)$
		\STATE $ \hat{\vv_t} = \max(\hat{\vv}_{t-1}, \hat{\vv}_t)$ (If using Amsgrad \citep{reddi2019convergence})
		\STATE $\theta_{t} = (1 - \alpha_t\gamma) \theta_{t-1} - \alpha_t\frac{\hat{\vm}_t}{(\hat{\vv}_t + \epsilon)^p}$
		\IF{$T \geq T_{\mathrm{avg}}$}
		\STATE $n_{\mathrm{models}} = n_{\mathrm{models}} + 1$
		\STATE $\theta_{\mathrm{avg}} = \frac{\theta_{\mathrm{avg}}\cdot n_{\mathrm{models}} + \theta_t}{n_{\mathrm{models}} + 1}$
		\ELSE
		\STATE $\theta_{\mathrm{avg}} = \theta_t$
		\ENDIF
		\ENDFOR
		\STATE \textbf{return} $\Tilde{\theta} = \theta_{\mathrm{avg}}$
	\end{algorithmic}
\end{algorithm}

\section{Derivations}
\label{sec:derivations}
In this section we give the proofs for our theorems.
\subsection{Proof of Theorem $1$}
\label{sec:theorem1proof}
The basic idea, is that since
\begin{equation}
	\mathbb{E}||\vw||_{2}^{2} = \sum_{i}^{p}\mathbb{E}(w_{i}^{2}) = \sum_{i}^{p}\bigg((\mathbb{E}(w_{i}))^{2}+\mathbb{V}(w_{i})\bigg) 
\end{equation}
we expect $||\vw||_{2}$ to be approximately the square root of this. Our proof follows very closely from \cite{vershynin2018high} (p.51) except that the variables we consider are not zero-mean or unit-variance. We sketch the proof below:
\paragraph{Proof Sketch:}
To show that Theorem 1 is indeed true with high probability, we consider the centered, unit variance version of the random variables i.e $X_{i}=(\tilde{X}_{i}-\mu_{i})/\sigma_{i}$
\paragraph{Lemma 1}\textit{(Bernstein's inequality)}: Let $\{X_{1},..X_{n}\}$ be independent, zero mean, sub-exponential random variables. Then for every $ t\geq 0$, we have
\begin{equation}
	\mathbb{P}\bigg \{ \bigg|\frac{1}{N}\sum_{i}\sum_{i}^{N}X_{i}\bigg| \geq t \bigg \} \leq 2\exp\bigg\{-c\text{min}\bigg(\frac{t^{2}}{K^{2}},\frac{t}{K}\bigg)N\bigg\}
\end{equation}
where $K = \argmax_{i}||X_{i}||_{\phi_{1}}$, and $||X||_{\phi_{i}} = \inf \{ t>0: \mathbb{E}\exp{|X|/t}\leq 2$.

\begin{proof}
	The proof is standard and can be found in \cite{vershynin2018high} p.45, essentially we multiply both sides of the inequality by $\lambda$, exponentiate, use Markov's inequality and independence assumption. Then we bound the MGF of each $X_{i}$ and optimise for $\lambda$
\end{proof}

\begin{proof}[Proof of Theorem $1$]
Let $X = (X_{1},...,X_{n}) \in \mathbb{R}^{P}$ be a random vector with independent sub-gaussian coordinates $X_{i}$, that satisfy $\mathbb{E}X_{i}^{2}=1$.
We then apply Berstein's deviation inequality (Lemma 1) for the normalized sum of independent, mean zero variables
\begin{equation}
	\frac{1}{P}||X||^{2}_{2}-1 = \frac{1}{n}\sum_{i}^{P}(X_{i}^{2}-1)
\end{equation}
Since  $X_{i}$ is sub-gaussian $X_{i}^{2}-1$ is sub-exponential and by centering and the boundedness of the MGF $||X_{i}^{2}-1||_{\phi_{1}}\leq CK^{2}$ hence assuming $K\geq 1$
\begin{equation}
	\mathbb{P}\bigg\{ \bigg|\frac{1}{P}||X||^{2}_{2}-1\bigg| \geq u \bigg\} \leq 2\exp \bigg( -\frac{cP}{K^{4}}\min(u^{2},u)\bigg)
\end{equation}
Then using $|z-1| \geq \delta$ implies $|z^{2}-1|\geq \max(\delta,\delta^{2})$
\begin{equation}
	\label{eq:keyequation}
	\begin{aligned}
		& \mathbb{P}\{\bigg|\frac{1}{\sqrt{P}}||X||_{2}-1\bigg|\geq \delta \bigg\} \\
		& \leq \mathbb{P}\{\bigg|\frac{1}{P}||X||^{2}_{2}-1\bigg|\geq \max(\delta,\delta^{2}) \bigg\}\\
		& \leq 2 \exp\left\{-\frac{cP}{K^{4}\delta^{2}}\right\}\\
	\end{aligned}
\end{equation}
changing variables to $t=\delta\sqrt{P}$ we obtain
\begin{equation}
	\mathbb{P}\{||X||_{2}-\sqrt{P}| \geq t\} \leq 2 \exp\left\{-\frac{ct^{2}}{K^{4}}\right\}
\end{equation}
for all $t\geq 0$
Our proof follows by noting that the significance of the $1$ in \eqref{eq:keyequation} is simply the mean of the square and hence replacing it by the mean squared plus variance is sufficient to obtain the result. To that end, with $\mLambda=\text{diag}\left(\lambda_1, \ldots, \lambda_P\right)$, we have 
 \begin{align}\label{eq:wn_expression_thm1}
    \vw_n = \sum_{i=1}^P(1-\alpha\mLambda)^n \vw_0 + \alpha\sum_{i=0}^{n-1} (1-\alpha\mLambda)^{n-i-1}\vepsilon_i
\end{align}
and then summing gives \begin{align}
    \vw_{avg} &= \frac{1 - (1 - \alpha\mLambda)^n}{\alpha n} \mLambda^{-1}(1-\alpha\mLambda)\vw_0 + \sum_{i=0}^{n-1} \frac{1 - (1-\alpha\mLambda)^{n-i}}{n}\mLambda^{-1}\vepsilon_i \label{eq:wavg_expression_thm1}.
    \end{align}
With all the $\vepsilon_i$ being i.i.d. $\mathcal{N}(0, \sigma^2B^{-1}I)$, we need simply to compute the sums \begin{align}
      \sum_{i=1}^{n-1} \alpha^2(1-\alpha\mLambda)^{2(n-i-1)} &= \alpha^2(1 - (1 - \alpha\mLambda)^{2n})\left(1 - (1-\alpha\mLambda)^2\right)^{-1}
\end{align}
and similarly \begin{align}
    \sum_{i=0}^{n-1} \left(\frac{1 - (1-\alpha\mLambda)^{n-i}}{n}\mLambda^{-1}\right)^2 = \frac{\mLambda^{-2}}{ n^2}\left(n - \frac{2(1 - (1-\alpha\mLambda)^n)}{\alpha}\mLambda^{-1} + \left(1 - (1-\alpha\mLambda)^{2n}\right)\left(1 - (1-\alpha\mLambda)^2\right)^{-1}\right).
\end{align}
Now assuming $\alpha\lambda_i \ll 1$ for all $i=1,2\ldots, P$, and taking $n\rightarrow\infty$, we find \begin{align}
      \sum_{i=1}^{n-1} (1-\alpha\mLambda)^{2(n-i-1)} \sim \alpha^2\left(1 - (1-\alpha\mLambda)^2\right)^{-1} = \alpha\left(2\mLambda - \alpha\mLambda^2\right)^{-1} 
\end{align}
and similarly \begin{align}
    \sum_{i=0}^{n-1} \left(\frac{1 - (1-\alpha\mLambda)^{n-i}}{n}\mLambda^{-1}\right)^2 \sim \frac{1}{ n}\mLambda^{-2}.
\end{align}
Thus it follows that \begin{align}
    \mathbb{E}w_{n, i}^2 = (1-\alpha\lambda_i)^n \sim e^{-2\alpha\lambda_i},& ~~ Var(w_{n, i}) \sim \frac{\sigma^2}{B}\frac{\alpha}{2\lambda_i(1 - \alpha\lambda_i)}
\end{align}
and \begin{align}
    \mathbb{E}w_{avg, i}^2 \sim \frac{w_{0,i}^2}{\lambda_i^2\alpha^2 n^2}, ~~~ Var(w_{avg, i}) = \frac{\sigma^2}{B}\frac{1}{n\lambda_i^2}
\end{align}
which completes the proof.

\end{proof}


\subsection{Proof of Theorem $2$}
Recall Theorem $2$ from the main text:

\begin{customthm}{2}
Let $\vw_n$ and $\vw_{avg}$ be defined as in Theorem $1$ and let the gradient noise be given by the covariance structure in Equation $6$ in the main text. Assume that the kernel function $k$ is such that $k'(-x^2)$ and $x^2k''(-x^2)$ decay as $x\rightarrow\infty$, and define $\sigma^2B^{-1} = k'(0)$ Assume further that $P\gg \log n$. Let $\delta=o(P^{1/2})$. Then $\vw_n$ and $\vw_{avg}$ are multivariate Gaussian random variables and, with probability which approaches unity as $P, n\rightarrow\infty$ the iterates $\vw_t$ are all mutually at least $\delta$ apart and 
\begin{align}
    &\mathbb{E}w_{n,i} \sim e^{-\alpha\lambda_i n}w_{0,i} , \\
    &\frac{1}{P}\Tr Cov(\vw_n) \sim \frac{\alpha\sigma^2}{B}\left\langle\frac{1}{\lambda(2-\alpha\lambda)}\right\rangle ,\\
    &\mathbb{E}w_{avg, i} \sim \frac{1-\alpha\lambda_i}{\alpha\lambda_i n}w_{0,i},\\
    &\frac{1}{P}\Tr Cov(\vw_{avg}) \leq \frac{\sigma^{2}}{Bn}\left\langle\frac{1}{ \lambda^2 }\right\rangle +  \mathcal{O}(1)\Bigg(k'(-\frac{\delta^2}{2}) + P^{-1}\delta^2k''(-\frac{\delta^2}{2})\Bigg).
\end{align}
\end{customthm}
The proof of this theorem relies on several intermediate results which we now state and prove.

\begin{lemma}\label{lemma:balls_lemma}
Take any $\vx_0,\ldots, \vx_{n-1}\in\mathbb{R}^{P}$ let $\mX\sim\mathcal{N}(0, \sigma^2I)$. Consider $P\rightarrow\infty$ with $P\gg \log{n}$ and let $\delta >0$ be $o(P^{\frac{1}{2}})$ (note that $\delta$ and $n$ need not diverge with $P$, but they can). Define \begin{align*}B_i = \{\vx\in\mathbb{R}^P \mid ||\vx-\vx_i|| < \delta \},\end{align*}
then as $P \rightarrow \infty$ \begin{align}\mathbb{P}\left(\mX \in \bigcup_i B_i\right) \rightarrow 0\end{align}
and moreover as $P, n\rightarrow\infty$ \begin{align}\label{eq:balls_lemma_precise}
    n\mathbb{P}\left(\mX \in \bigcup_i B_i\right) \rightarrow 0.
\end{align}
\end{lemma}
\begin{proof}
With the Euclidean volume measure, we have \begin{align*}
    Vol\left(\bigcup_i B_i\right) \leq n V_P \delta^P = V_P (\delta n^{1/P})^{P}
\end{align*}
where $V_P$ is the volume of the unit sphere in $P$ dimensions. Therefore \begin{align}
\mathbb{P}\left(\mX \in \bigcup_i B_i\right) & \leq \frac{1}{(2\pi \sigma^2)^{\frac{P}{2}}}\frac{2\pi^{\frac{P}{2}}}{\Gamma(\frac{P}{2})}\int_0^{\delta n^{\frac{1}{P}}}dr ~ e^{-\frac{r^2}{2\sigma^2}} r^{P-1}\notag\\
&= \frac{2}{\Gamma(\frac{P}{2})}\int_0^{\frac{\delta n^{\frac{1}{P}}}{\sqrt{2}\sigma}} dr ~ e^{-r^2} r^{P-1}\notag\\
& = \frac{1}{\Gamma(\frac{P}{2})} \int_0^{\frac{\delta n^{\frac{2}{P}}}{2\sigma^2}} dr ~ e^{-r} r^{\frac{P}{2} - 1}\notag\\
& = \frac{1}{\Gamma(\frac{P}{2})} \gamma\left(\frac{P}{2}; \frac{n^{\frac{2}{P}}\delta^2}{2\sigma^2}\right)\label{eq:ball_prob_gamma}
\end{align}
where $\gamma$ is the lower incomplete gamma function. Since $P\gg \log{n}$ and $\delta = o(P^{\frac{1}{2}})$, it follows that \begin{align*}
    \frac{n^{\frac{2}{P}}\delta^2}{2\sigma^2} = o(P)
\end{align*}
and so Lemma \ref{lemma:incomp_gamma} can be applied to yield the result, recalling that $n\ll e^{P}$.
\end{proof}

\begin{lemma}\label{lemma:incomp_gamma}
Define the function \begin{align}
    r(a; x) = \frac{\gamma(a; x)}{\Gamma(a)},
\end{align}
where $\gamma$ is the lower incomplete gamma function. Assume that $a\ll x$, where $x$ may or may not diverge with $a$, then as $a\rightarrow\infty$, $r(a; x)\rightarrow 0$, and more precisely \begin{align}
    r(a; x) \sim \frac{1}{\sqrtsign{2\pi}} \exp\left(-x + a\log{x} - a - a\log{a} - \frac{1}{2}\log{a}\right).
\end{align}
\end{lemma}
\begin{proof}
    We have $\gamma(a; x) = a^{-1}x^a \,_1F_1(a; 1+a; -x)$, where $\,_1F_1$ is the confluent hypergeometric function of the first kind \citep{andrews_askey_roy_1999}. Then \begin{align}
    r(a; x) &= \frac{a^{-1} x^a \,_1F_1(a; 1+a; -x)}{\Gamma(a)}=\frac{a^{-1} x^a \Gamma(a+1)}{\Gamma(a)^2}\int_0^1 e^{xt} t^{a-1} dt\label{eq:r_int_form}
\end{align}
where we have used a result of \citet{abramowitz1988handbook}. The integral in (\ref{eq:r_int_form}) can be evaluated asymptotically in the limit $x\rightarrow\infty$ with $x \ll a$. Writing the integrand as $e^{xt + (a-1)\log{t}}$ it is plainly seen to have no saddle points in $[0, 1]$ given the condition $x\ll a$. The leading order term therefore originates at the right edge $t=1$. A simple application of Laplace's method leads to \begin{align*}
    r(a; x) & \sim  \frac{a^{-1} x^a \Gamma(a+1) e^{-x}}{\Gamma(a)^2(a- 1 -x)}\\
    & \sim \frac{ x^a  e^{-x}}{a\Gamma(a)}\\
    & \sim \frac{ x^a  e^{-x}}{a \sqrtsign{2\pi a^{-1}} (ae^{-1})^a}\\
    & = \frac{1}{\sqrtsign{2\pi}} \exp\left(-x + a\log{x} - a - a\log{a} - \frac{1}{2}\log{a}\right)
\end{align*}
where the penultimate line makes uses of Stirling's approximation \citep{andrews_askey_roy_1999}. Since $a\gg x$, \begin{align*}
    -x + a\log{x} - a - a\log{a} - \frac{1}{2}\log{a} \sim -a\log{a}\rightarrow-\infty 
\end{align*}
which completes the proof.
\end{proof}

\begin{lemma}\label{lemma:prob_far_apart}
Let $\mX_1,\ldots, \mX_n$ be a sequence of multivariate Gaussian random variables in $\mathbb{R}^P$ with \begin{align*}
    \mX_i \mid \{\mX_j \mid 0< j < i\} \sim \mathcal{N}(0, \sigma^2I)
\end{align*}
for all $1\leq i \leq n$. Let $\mX_0$ be any deterministic element of $\mathbb{R}^P.$ Define the events \begin{align*}A_m(\delta) = \{||\mX_i - \mX_j||_2 > \delta \mid 0\leq i < j \leq n\}.\end{align*} Let $n, P$ and $\delta$ satisfy the constraints in the statement of Lemma \ref{lemma:balls_lemma}. Then $\mathbb{P}(A_n(\delta))\rightarrow 1$ as $P\rightarrow\infty$.
\end{lemma}
\begin{proof}
Let us use the definitions of $B_i$ from Lemma \ref{lemma:balls_lemma} (with $\mX_i$ replacing $\vx_i$ and $\mX$ in the obvious way). Since $A_i(\delta) \subset A_{i-1}(\delta)$, the chain rule of probability gives \begin{align}
    \mathbb{P}(A_n(\delta))& = \mathbb{P}\left(\bigcap_{i\leq n}A_i(\delta)\right) = \mathbb{P}(A_1(\delta))\prod_{i=1}^{n-1} \mathbb{P}(A_i \mid A_{i-1})\notag
\end{align}
but \begin{align*}\mathbb{P}(A_i(\delta) \mid A_{i-1}(\delta))= 1 - \mathbb{P}\left(\mX_i\in\bigcup_{j< i} B_j\right)\end{align*} and so (\ref{eq:balls_lemma_precise}) gives the result.
\end{proof}

\begin{lemma}\label{lemma:gp_covar_trace}
Recall the covariance structure from the main text Equation 6
\begin{align}
\nonumber
    & Cov(\epsilon_i(\vw), \epsilon_j(\vw') )  = (w_i - w'_i)(w'_j - w_j) k''\left(-\frac{1}{2}||\vec{w} - \vec{w}'||_2^2\right) + \delta_{ij}k'\left(-\frac{1}{2}||\vec{w} - \vec{w}'||_2^2\right).
\end{align}
Take any $a_i\in\mathbb{R}$ and define $\bar{\vepsilon} = \sum_{i=1}^n a_i \vepsilon_i$. Then 
\begin{align}\label{eq:gp_trace_cov}
    \Tr~Cov(\bar{\vepsilon}) &= k'(0)P\sum_{i=1}^na_i^2 + 2P\sum_{1\leq i<j\leq n}a_ia_j\Bigg[k'(-\frac{d_{ij}^2}{2})+ P^{-1}k''(-\frac{d_{ij}^2}{2})d_{ij}^2\Bigg]
\end{align}
where we define $d_{ij} = ||\vw_i - \vw_j||_2$.
\end{lemma}
\begin{proof}
Each of the $\vepsilon_i$ is Gaussian distributed with covariance matrix $Cov(\vepsilon_i)$ given by Equation $6$ and the covariance between different gradients $Cov(\vepsilon_i, \vepsilon_j)$ is similarly given by Equation $6$. By standard multivariate Gaussian properties
\begin{align}
   Cov(\bar{\vepsilon}) &= \sum_{i=1}^na_i^2~Cov(\vepsilon_i)+ \sum_{i\neq j}a_ia_j Cov(\vepsilon_i, \vepsilon_j),
\end{align}
then taking the trace
\begin{align}
    \Tr~Cov(\bar{\vepsilon}) &= \sum_{i=1}^na_i^2\Tr(Cov(\vepsilon_i))+ 2\sum_{1\leq i<j\leq n}a_ia_j \Tr(Cov(\vepsilon_i, \vepsilon_j)).
    \end{align}
    Using the covariance structure from Equation $6$ in the main text gives \begin{align}
\Tr~Cov(\bar{\vepsilon}) = k'(0)\sum_{i=1}^na_i^2 \Tr{I} + 2\sum_{1\leq i<j\leq n}a_ia_j\Bigg[&k'(-\frac{d_{ij}^2}{2})\Tr{I}\notag\\
&+ k''(-\frac{d_{ij}^2}{2})\Tr(\vw_i - \vw_j)(\vw_j - \vw_i)^T\Bigg]
    \end{align}
from which the result follows.
\end{proof}

\begin{proof}[Proof of Theorem $2$]
We will prove the result in the case $\lambda_i = \lambda ~\forall i$ for the sake of clarity. The same reasoning can be repeated in the more general case; where one gets $P^{-1} f(\lambda) Tr I$ below, one need only replace it with $\langle f(\lambda)\rangle$. We will also vacuously replace $\sigma^2B^{-1}$ with $\sigma^2$ to save on notation.
For weight iterates $\vw_i$, we have the recurrence \begin{align*}
    \vw_i = (1-\alpha\lambda)\vw_{i-1} + \alpha\vepsilon(\vw_{i-1})
\end{align*}
which leads to \begin{align}\label{eq:wn_expression}
    \vw_n = (1-\alpha\lambda)^n \vw_0 + \alpha\sum_{i=0}^{n-1} (1-\alpha\lambda_i)^{n-i-1}\vepsilon(\vw_i)
\end{align}
and then \begin{align}
    \vw_{avg} &= \frac{1 - (1 - \alpha\lambda)^n}{\alpha\lambda n} (1-\alpha\lambda)\vw_0 + \sum_{i=0}^{n-1}\vepsilon(\vw_i) \frac{1 - (1-\alpha\lambda)^{n-i}}{\lambda n}\label{eq:wavg_expression}.
\end{align}
Now define \begin{align*}
    a_i = \alpha(1-\alpha\lambda)^{n-1-i}, ~~~ \bar{a}_i = \frac{1 - (1-\alpha\lambda)^{n-i}}{\lambda n}.
\end{align*}
Next we will apply Lemma \ref{lemma:gp_covar_trace} and utilise Lemma \ref{lemma:prob_far_apart} to bound the variance of $\vw_{avg}$ and $\vw_n$. We first gather the following facts (which were also computed and used in the proof of Theorem $1$:
\begin{align}
    \sum_{i=1}^{n-1} a_i^2 &= \frac{\alpha^2(1 - (1 - \alpha\lambda)^{2n})}{1 - (1-\alpha\lambda)^2}\\
    \sum_{i<j}a_ia_j &= \frac{\alpha}{\lambda}\left(\frac{1 - (1-\alpha\lambda)^n}{\alpha\lambda} - \frac{1 - (1-\alpha\lambda)^{2n}}{1 - (1-\alpha\lambda)^2}\right).
\end{align}
The sum of squares for the $\bar{a}_i$ is simple to obtain similarly \begin{align}
    \sum_{i=0}^{n-1} \bar{a}_i^2 = \frac{1}{\lambda^2 n^2}\left(n - \frac{2(1 - (1-\alpha\lambda)^n)}{\alpha\lambda} + \frac{1 - (1-\alpha\lambda)^{2n}}{1 - (1-\alpha\lambda)^2}\right).
\end{align}
We now use the assumption that $0 < \alpha\lambda < 1$ (required for the convergence of gradient descent) which gives, as $n\rightarrow\infty$, \begin{align}
     \sum_{i=1}^{n-1} a_i^2 & \sim \frac{\alpha^2}{1 - (1 -\alpha\lambda)^2}\label{eq:ai_sq_sum}\\
     \sum_{i<j}a_ia_j & \sim  \frac{\alpha}{\lambda}\left(\frac{1}{\alpha\lambda} - \frac{1}{1 - (1-\alpha\lambda)^2}\right)\label{eq:aiaj_sum}\\
      \sum_{i=1}^{n-1} \bar{a}_i^2 & \sim \frac{1}{\lambda^2 n}\label{eq:abari_sq_sum}
\end{align}
Summing $\sum_{i< j}\bar{a}_i\bar{a}_j$ explicitly is possible but unhelpfully complicated. Instead, some elementary bounds give \begin{align*}
    \sum_{i< j}\bar{a}_i\bar{a}_j &\leq \left(\sum_{i=0}^{n-1}\bar{a}_i\right)^2 = \frac{1}{\lambda^2n^2}\left(n - \frac{1 - (1-\alpha\lambda)^n}{\alpha\lambda}\right)^2 \sim\frac{1}{\lambda^2}
\end{align*}
and \begin{align*}
    \sum_{i< j}\bar{a}_i\bar{a}_j &\geq \sum_{i<j}\left(\frac{1 - (1-\alpha\lambda)^{n-1}}{\lambda n}\right)^2 \sim \frac{1}{2\lambda^2}
\end{align*}
so in particular $\sum_{i < j}\bar{a}_i\bar{a}_j = \mathcal{O}(1)$. Now let $P, n, \delta$ satisfy the conditions of Lemma \ref{lemma:balls_lemma}, and define the events $A_n(\delta)$ as in Lemma \ref{lemma:prob_far_apart} using $\vepsilon_i$ in place of $\mX_i$. Further, choose $\delta$ large enough so that $k'(-\frac{x^2}{2})$ and $x^2k''(-\frac{x^2}{2})$ are decreasing for $x>\delta$. Define $k'(0) = \sigma^2$. Lemma \ref{lemma:gp_covar_trace} gives \begin{align}
   \frac{1}{P}\Tr Cov(\vw_n) \mid A_n(\delta) &\leq \sigma^2\sum_{i=1}^n a_i^2 + 2\sum_{i<j} a_ia_j\Bigg(k'(-\frac{\delta^2}{2}) + P^{-1}\delta^2k''(-\frac{\delta^2}{2})\Bigg)\label{eq:trace_wn_bound}
\end{align}
where note that we have only upper-bounded the second term in (\ref{eq:trace_wn_bound}), so using  (\ref{eq:ai_sq_sum}) and (\ref{eq:aiaj_sum}) and taking $\delta$ large enough we get \begin{align}
    \frac{1}{P}\Tr Cov(\vw_n) \mid A_n(\delta) = \frac{\sigma^2\alpha^2}{1 - (1-\alpha\lambda)^2} + o(1).
\end{align}
Turning now to $\vw_{avg}$ we similarly obtain \begin{align}
    \frac{1}{P}\Tr Cov(\vw_{avg}) \mid A_n(\delta) \leq \frac{\sigma^2}{n}\frac{1}{\lambda^2}+ \mathcal{O}(1)\Bigg(k'(-\frac{\delta^2}{2}) + P^{-1}\delta^2k''(-\frac{\delta^2}{2})\Bigg)
\end{align}
and, as before, taking $\delta$ large enough we can obtain \begin{align}
    \frac{1}{P}\Tr Cov(\vw_{avg}) \mid A_n(\delta) = o(1).
\end{align}
Finally recalling (\ref{eq:wn_expression}) and (\ref{eq:wavg_expression}) and writing $(1-\alpha\lambda)^n = e^{-\alpha\lambda n}+ o(1)$ for large $n$, we obtain the result.
\end{proof}

\begin{customthm}{3}
Let $\vw_n$ and $\vw_{avg}$ be defined as in Theorem $2$ and let the gradient noise be given by the basic covariance structure in Equation $6$ in the main text.
Let the kernel function be of the form \begin{align*}
    k(\vw, \vw') = k\left(-\frac{||\vw - \vw'||_2^2}{||\vw||_2^2}\right)
\end{align*}and assume that the kernel function $k$ is such that $k'(-x^2)$ and $x^2k''(-x^2)$ decay as $x\rightarrow\infty$, and define $\sigma^2B^{-1} = k'(0)$ Assume further that $P\gg \log n$. Then the result of Theorem $2$ holds.
\end{customthm}

\begin{proof}
The proof is much the same as that of Theorem $2$. The expression analogous to Equation $6$ in the main text is different in the following two ways:

\begin{enumerate}
    \item All terms are divided by positive powers of $||w||_2$.
    \item There are extra terms arising from derivatives applied to $||\vw||_2^{-1}$, which give rise to factors of the form $\vw_i^T(\vw_j - \vw_j)$.
\end{enumerate}
Extending the proof of Theorem $2$ to this case requires the following two observations. Firstly, terms of the form $\vw_i^T(\vw_j - \vw_j)$ can be easily bounded as $|\vw_i^T(\vw_j - \vw_j)|\leq ||\vw||_2 d_{ij}$. Secondly the terms of the form $||\vw||_2^{-r}$ for $r \geq 1$ will cause no problems so long as they can be uniformly bounded away from $0$ as $P, n\rightarrow\infty$. This can be established with high probability as a trivial extension of Lemma \ref{lemma:prob_far_apart} by introducing an extra point $\mX_0$, say, at the origin.
\end{proof}

\begin{corollary}\label{cor:strided}
Let $\vw_{avg}$ now be a strided iterate average with stride $\kappa$, i.e. \begin{align}
    \vw_{avg} = \frac{\kappa}{n}\sum_{i=1}^{\lfloor n/\kappa \rfloor} \vw_i.
\end{align}
Then, under the same conditions as Theorem $2$ (or Theorems $3$ or $4$)
\begin{align}
    &\mathbb{E}w_{avg, i} = \frac{\kappa(1-\alpha\lambda_i)^{\kappa}}{n(1 - (1-\alpha\lambda_i)^{\kappa} )} (1 + o(1))w_{0,i},\\
    &\frac{1}{P}\Tr Cov(\vw_{avg}) \leq \frac{\sigma^2\alpha^2\kappa}{Bn}\left\langle\frac{1}{\left(1 - (1-\alpha\lambda)^{\kappa}\right)^2}\frac{1 - (1-\alpha\lambda)^{2\kappa}}{1 - (1-\alpha\lambda)^{2}}\right\rangle+  \mathcal{O}(1)\Bigg(k'(-\frac{\delta^2}{2}) + P^{-1}\delta^2k''(-\frac{\delta^2}{2})\Bigg)\label{eq:corr_ia_gp_covar}
\end{align}
where the constant $\mathcal{O}(1)$ coefficient of the second term in (\ref{eq:corr_ia_gp_covar}) is independent of $\kappa$.
\end{corollary}
\begin{proof}
The proof is just as in Theorem $2$ (or Theorems $3$ or $4$), differing only in the values of the $\bar{a}_i$. Indeed, a little thought reveals that the generalisation of $\bar{a}_i$ to the case $\kappa > 1$ is \begin{align}
    \bar{a}_i = \frac{\alpha\kappa}{n}(1 - \alpha\lambda)^{\kappa\left(1 + \lfloor\frac{i}{\kappa}\rfloor\right) - 1 - i} \frac{1 - (1-\alpha\lambda)^{\kappa\left(\lfloor\frac{n}{\kappa}\rfloor - \lfloor\frac{i}{\kappa}\rfloor\right)}}{1 - (1-\alpha\lambda)^{\kappa}}.
\end{align}
Note that $\kappa \left\lfloor\frac{i}{\kappa}\right\rfloor - i $ is just the (negative) remainder after division of $i$ by $\kappa$. 
Then for large $n$ \begin{align*}
    \sum_{i} \bar{a}_i^2 &\sim \frac{\alpha^2\kappa^2}{n^2}\frac{(1-\alpha\lambda)^{2(\kappa - 1)}}{\left(1 - (1-\alpha\lambda)^{\kappa}\right)^2} \left\lfloor\frac{n}{\kappa}\right\rfloor \sum_{i=0}^{\kappa-1}(1-\alpha\lambda)^{-2i}\\
     &\leq \frac{\alpha^2\kappa}{n}\frac{(1-\alpha\lambda)^{2(\kappa - 1)}}{\left(1 - (1-\alpha\lambda)^{\kappa}\right)^2} \sum_{i=0}^{\kappa-1}(1-\alpha\lambda)^{-2i}\\
      &= \frac{\alpha^2\kappa}{n}\frac{(1-\alpha\lambda)^{2(\kappa - 1)}}{\left(1 - (1-\alpha\lambda)^{\kappa}\right)^2}\frac{1 - (1-\alpha\lambda)^{-2\kappa}}{1 - (1-\alpha\lambda)^{-2}}\\
    &= \frac{\alpha^2\kappa}{n}\frac{1}{\left(1 - (1-\alpha\lambda)^{\kappa}\right)^2}\frac{1 - (1-\alpha\lambda)^{2\kappa}}{1 - (1-\alpha\lambda)^{2}}.
\end{align*}
and similarly \begin{align}
    \sum_{i< j} \bar{a}_i \bar{a}_j &\sim \frac{\alpha^2\kappa^2}{n^2}\frac{(1-\alpha\lambda)^{2(\kappa-1)}}{(1 - (1-\alpha\lambda)^{\kappa})^2}\sum_{i<j} (1-\alpha\lambda)^{\kappa\lfloor i/\kappa\rfloor - i + \kappa\lfloor j/\kappa\rfloor - j}\\
    &\sim \frac{\alpha^2\kappa^2}{n^2}\frac{(1-\alpha\lambda)^{2(\kappa-1)}}{(1 - (1-\alpha\lambda)^{\kappa})^2}\sum_j (1-\alpha\lambda)^{\kappa\lfloor j/\kappa\rfloor - j}\left\lfloor\frac{j}{\kappa}\right\rfloor\frac{1 - (1-\alpha\lambda)^{-\kappa}}{1 - (1-\alpha\lambda)^{-1}}\\
    &\sim  \frac{\alpha^2\kappa^2}{n^2}\frac{(1-\alpha\lambda)^{2(\kappa-1)}}{(1 - (1-\alpha\lambda)^{\kappa})^2}\left(\frac{1 - (1-\alpha\lambda)^{-\kappa}}{1 - (1-\alpha\lambda)^{-1}}\right)^2\sum_{j=0}^{\lfloor n/\kappa\rfloor}j \\
    &\sim  \frac{\alpha^2}{2}\frac{(1-\alpha\lambda)^{2(\kappa-1)}}{(1 - (1-\alpha\lambda)^{\kappa})^2}\left(\frac{1 - (1-\alpha\lambda)^{-\kappa}}{1 - (1-\alpha\lambda)^{-1}}\right)^2\\
    &=  \frac{\alpha^2}{2}\frac{(1-\alpha\lambda)^{-2}}{(1 - (1-\alpha\lambda)^{-1})^2}.
\end{align}
\end{proof}

\begin{customthm}{4}
Fix some $\zeta > 0$ and assume that $|\tilde{\lambda}_i^{(t)} - \lambda_i| < \zeta$ for all $t \geq n_0$, for some fixed $n_0(\zeta)$, with high probability. Use the update rule Equation $9$ in the main text. Assume that the $\lambda_i$ are bounded away from zero and $\min_i\lambda_i > \zeta$. Further assume $c(\gamma + \varepsilon + \zeta) < 1$, where $c$ is a constant independent of $\varepsilon, \zeta, \gamma$ and is defined in the proof. Let everything else be as in Theorem $2$. Then there exist constants $c_1, c_2, c_3, c_4>0$ such that, with high probability,
\begin{align}
    &|\mathbb{E}w_{n,i}-w^*_i| \leq e^{-\alpha(1 +\gamma -c(\varepsilon+\zeta))n}w_{0,i} + c_1(\varepsilon +\zeta + \gamma)\label{eq:thm4_proof1} \\
    &\left|\frac{1}{P}\Tr Cov(\vw_n)- \frac{\alpha\sigma^2}{B(2-\alpha)}\right| \leq   c_2(\varepsilon + \zeta + \gamma) + o(1),\label{eq:thm4_proof2}\\
    &|\mathbb{E}w_{avg, i} - w^*_i| \leq \frac{1-\alpha(1 +\gamma - c(\varepsilon + \zeta ))}{\alpha(1 + \gamma - c(\varepsilon+\zeta)) n} (1 + o(1))w_{0,i} + c_3(\varepsilon + \zeta + \gamma)\label{eq:thm4_proof3}\\
    &\left|\frac{1}{P}\Tr Cov(\vw_{avg}) - \frac{\sigma^{2}}{Bn} -  \mathcal{O}(1)\Bigg(k'(-\frac{\delta^2}{2}) + P^{-1}\delta^2k''(-\frac{\delta^2}{2})\Bigg)\right| \leq c_4 (\gamma, + \zeta + \epsilon).\label{eq:thm4_proof4}
\end{align}
\end{customthm}

\begin{proof}
We begin with the equivalent of (\ref{eq:wn_expression}) for update rule Equation $9$ in the main text: \begin{align}
    \vw_n = \prod_{i=0}^{n-1}\left(1-\alpha\gamma -\alpha \tilde{H}_i^{-1}\Lambda\right)\vw_0 &+ \sum_{i=}^{n-1}\alpha \tilde{H}_i^{-1}\Lambda \prod_{j=i+1}^{n-1} \left(1 - \alpha\gamma -\alpha\tilde{H}_j^{-1}\Lambda \right)\vw^* \notag\\
    & - \sum_{i=}^{n-1}\alpha \tilde{H}_i^{-1}\Lambda \left[\prod_{j=i+1}^{n-1} \left(1 - \alpha\gamma -\alpha\tilde{H}_j^{-1}\Lambda \right)\right]\vepsilon(\vw_i).
\end{align}
To make progress, we need the following bounds valid for all $t\geq n_0$
 \begin{align*}
     \frac{\lambda_i}{\tilde{\lambda}_i^{(t)}+ \varepsilon} = \frac{\lambda_i}{\lambda_i + \tilde{\lambda}_i^{(t)}  -\lambda_i + \varepsilon}<\frac{\lambda_i}{\lambda_i + \varepsilon - \zeta} < 1 + |\varepsilon-\zeta|\lambda_i^{-1}
\end{align*}

and \begin{align*}
     \frac{\lambda_i}{\tilde{\lambda}_i^{(t)}+ \varepsilon} = \frac{\lambda_i}{\lambda_i + \tilde{\lambda}_i^{(t)}  -\lambda_i + \varepsilon}>\frac{\lambda_i}{\lambda_i + \varepsilon + \zeta} > 1 - (\varepsilon + \zeta)\lambda_i^{-1}
\end{align*}
where the final inequality in each case can be derived from Taylor's theorem with Lagrange's form of the remainder. 

Since the $\lambda_i$ are bounded away from zero, we have established \begin{align}\label{eq:precondition_success_bound}
      \left|\frac{\lambda_i}{\tilde{\lambda}_i^{(t)}+ \varepsilon} - 1\right| < c(\varepsilon + \zeta)
\end{align}
where the constant $c= 1+(\min_j\{\lambda_j\})^{-1}$, say.
From this bound we can in turn obtain \begin{align}
 &1 - \alpha(\gamma + 1 + c(\varepsilon + \zeta))<   1 - \alpha(\gamma + \tilde{(\lambda}_i^{(t)} + \varepsilon)^{-1}\lambda_i) < 1 - \alpha(\gamma + 1 - c(\varepsilon + \zeta))\notag\\
 \implies &1 - \alpha( 1 + c(\varepsilon + \zeta + \gamma))<   1 - \alpha(\gamma + \tilde{(\lambda}_i^{(t)} + \varepsilon)^{-1}\lambda_i) < 1 - \alpha(1 - c(\varepsilon + \zeta + \gamma))
\end{align}
where the second line exploits the assumption $c(\gamma + \varepsilon + \zeta) < 1$ and our choice $c > 1$. Thus \begin{align}
    \sum_{t=0}^{n-1} \alpha \frac{\lambda_k}{\tilde{\lambda}_k^{(t)}} \prod_{j=t+1}^{n-1}\left(1-\alpha\gamma -\alpha(\tilde{\lambda}_k^{(j)} + \varepsilon)\lambda_k\right) &< \sum_{t=0}^{n-1} \alpha (1 + c(\varepsilon + \zeta))\left( 1 - \alpha(\gamma + 1 - c(\varepsilon + \zeta))\right)^{n-1-t}\notag\\
    &< 1 + c_1(\zeta + \varepsilon + \gamma)
\end{align}
where the second inequality follows, for large $n$, by summing the geometric series and again using Lagrange's form of the remainder in Taylor's theorem. $c_1$ is some constant, derived from $c$ that we need not determine explicitly. A complementary lower bound is obtained similarly (for large $n$). We have thus shown that \begin{align}
    |\mathbb{E}w_{n,i} - w^*_i| < c_1(\varepsilon + \zeta + \gamma) + \prod_{t=0}^{n-1}\left(1-\alpha\gamma -\alpha (\tilde{\lambda}_i^{(t)} + 
    \varepsilon)^{-1}\lambda_i\right)w_{0,i}.
\end{align}
Reusing the bound (\ref{eq:precondition_success_bound}) then yields (\ref{eq:thm4_proof1}). The remaining results, (\ref{eq:thm4_proof2})-(\ref{eq:thm4_proof4}) follow similarly using the same bounds and ideas as above, but applied to the corresponding steps from the proof of Theorem $2$.
\end{proof}

\section{Gadam and Lookahead}
\label{sec:lookahead}
As discussed, most related works improve generalisation of adaptive methods by combining them with SGD in some form. As an example representing the recent works claiming promising performances, \cite{chen2018closing} introduce an additional hyperparameter $p$,to ontrols the extent of adaptivity: for $p = \{\frac{1}{2},0\}$, we have fully adaptive Adam(W) or pure first-order SGD respectively and usually a $p$ falling between the extremes is taken. In addition to empirical comparisons, since our approach is orthogonal to these approaches, as an singular example, we propose \textbf{GadamX} that combines Gadam with Padam, where for simplicity we follow \cite{chen2018closing} to fix $p=\frac{1}{8}$ for the current work. We note that $p<1$ is regularly considered a heuristic to be used for an inaccurate curvature matrix \cite{martens2014new}, although the specific choice of $p=1/2$ has a principled derivation in terms of a regret bound \cite{duchi2011adaptive}. Previous works also use EMA in weight space to achieve optimisation and/or generalisation improvements: \cite{izmailov2018averaging} entertain EMA in SWA, although they conclude simple averaging is more competitive. Recently, \cite{zhang2019lookahead} proposes \textit{Lookahead} (LH), a plug-in optimiser that uses EMA on the slow weights to improve convergence and generalisation. Nonetheless, having argued the dominance of noise in the high-dimensional deep learning regime, we argue that simple averaging is more theoretically desirable \textit{for generalisation}. Following the identical analysis to the noisy quadratic with i.i.d noise, we consider the $1D$ case w.l.o.g and denote $\rho \in [0, 1]$ as the coefficient of decay, asymptotically the EMA point $\vw_{\mathrm{ema}}$ is governed by:
\small
\begin{equation}
	\mathcal{N}\bigg(\frac{(1-\rho)w_{0}(1-\alpha\lambda)^{n+1}[1-(\frac{\rho}{1-\alpha\lambda})^{n-1}]}{1-\alpha\lambda-\rho},\frac{1-\rho}{1+\rho}\frac{\alpha\sigma^{2}\kappa}{\lambda}\bigg) 
	\label{eq:ema}
\end{equation}
\normalsize
Where $\kappa = (1-(1-\alpha\lambda)^{n-2})$).
An alternative analysis of EMA arriving at similar result was done in \cite{zhang2019algorithmic}, but their emphasis of comparison is between the EMA and \textit{iterates} instead of EMA and the \textit{IA point} in our case. From equation \ref{eq:ema}, while the convergence in mean is less strongly affected, the noise is reduced by a factor of $\frac{1-\rho}{1+\rho}$. So whilst we reduce the noise possibly by a very large factor, it does not vanish asymptotically. Hence viewing EMA or IA as noise reduction schemes, we consider IA to be far more aggressive. Secondly, EMA implicitly assumes that more recent iterates are better, or otherwise more important, than the previous iterates. While justified initially (partially explaining LH's efficacy in accelerating optimisation), it is less so in the late stage of training. We nonetheless believe LH could be of great combinable value, and include a preliminary discussion in App. \ref{sec:lookahead}. 

\emph{Lookahead} \citep{zhang2019lookahead} is a very recent attempt that also features weight space averaging in order to achieve optimisation and generalisation benefits. However, instead of using simple averaging in our proposed algorithms, Lookahead maintains different update rules for the \textit{fast} and \textit{slow} weights, and uses exponentially moving average to update the parameters. In this section, we both comment on the key theoretical differences between Gadam and Lookahead and make some preliminary practical comparisons. We also offer an attempt to bring together the \textit{optimisation} benefit of Lookahead and the \textit{generalisation} benefit of Gadam, with promising preliminary results.

\subsubsection{Major Differences between Gadam and Lookahead}
\paragraph{Averaging Method} Lookahead opts for a more complicated averaging scheme: they determine the 'fast'- and 'slow'-
varying weights during optimisation, and maintains an EMA to average the weight. On the other hand, Gadam uses a more straightforward simple average. As we discussed in the main text, EMA is more theoretically justified during the initial rather than later stage of training. This can also be argued from a Bayesian viewpoint following \cite{maddox2019simple}, who argued that iterates are simply the draws from the posterior predictive distribution of the neural network, where as averaging leads to a rough estimation of its posterior mean. It is apparent that if the draws from this distribution are \textit{equally} good (which is likely to be the case if we start averaging only if validation metrics stop improving), assigning the iterates with an exponential weight just based on when they are drawn constitutes a rather arbitrary prior in Bayesian sense.

\paragraph{Averaging Frequency} Lookahead averages every iteration whereas in Gadam, while possible to do so as well, by default averages much less frequently. We detail our rationale for this in Appendix \ref{sec:avgfreq}. 

\paragraph{Starting Point of Averaging} 
While Lookahead starts averaging at the beginning of the training, Gadam starts averaging either from a pre-set starting point or an automatic trigger (for GadamAuto). While authors of Lookahead \cite{zhang2019lookahead} argue that starting averaging eliminates the hyperparameter on when to start averaging, it is worth noting that Lookahead also introduces two additional hyperparameters $\alpha$ and $k$, which are non-trivially determined from grid search (although the authors argue that the final result is not very sensitive to them). 

We believe the difference here is caused by the different design philosophies of Gadam and Lookahead: by using EMA and starting averaging from the beginning, Lookahead benefits from faster convergence and some generalisation improvement whereas in Gadam, since the averages of iterates are not used during training to promote independece between iterates, Gadam does not additionally accelerate optimisation but, by our theory, should generalise better. As we will see in the next section, this theoretical insight is validated by the experiments and leads to combinable benefits.

\subsubsection*{Empirical Comparison}
We make some empirical evaluations on CIFAR-100 data-set with different network architectures, and we use different base optimiser for Lookahead. For all experiments, we use the author-recommended default values of $k = 5$ (number of lookahead steps) and $\alpha = 0.5$. We focus on the combination of Lookahead and adaptive optimisers, as this is the key focus of this paper, although we do include results with Lookahead with SGD as the base optimiser.

\begin{figure}[h]
	\centering
	\begin{subfigure}{0.49\linewidth}
		\includegraphics[width=1\linewidth]{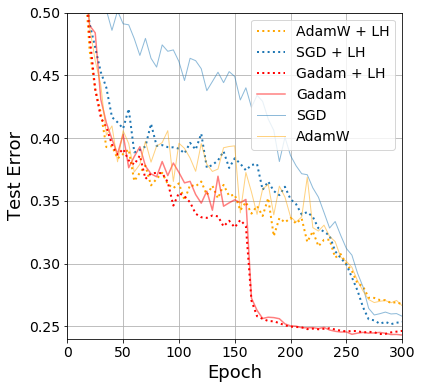}
		\caption{VGG-16}
	\end{subfigure}
	\begin{subfigure}{0.49\linewidth}
		\includegraphics[width=1\linewidth]{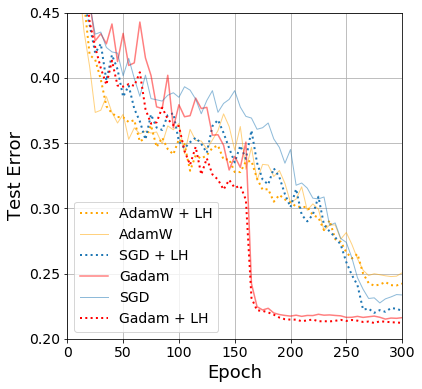}
		\caption{PRN-110}
	\end{subfigure}
	
	\caption{Test accuracy of Lookahead in CIFAR-100 against number of epochs. }\label{fig:lookaheadvgg}
\end{figure}

We first test AdamW and SGD with and without Lookahead and the results are in Figure \ref{fig:lookaheadvgg}. Whilst SGD + LH outperforms SGD in final test accuracy by a rather significant margin in both architectures, Lookahead does not always lead to better final test accuracy in AdamW (although it does improve the convergence speed and reduce fluctuations in test error during training, which is unsurprising as EMA shares similar characteristics with IA in reducing sensitivity to gradient noise). On the other hand, it is clear that Gadam delivers both more significant and more consistent improvements over AdamW, both here and in the rest of the paper.


Nonetheless, we believe that Lookahead, being an easy-to-use plug-in optimiser that clearly improves convergence speed, offers significant combinable potential with Gadam, which focuses on generalisation. Indeed, by using Lookahead \textit{before} the 161st epoch where we start IA, and switching to IA \textit{after} the starting point, we successfully combine Gadam and LH into a new optimiser which we term Gadam + LH. With reference to Figure \ref{fig:lookaheadvgg}, in VGG-16, Gadam + LH both converges at the fastest speed in all the optimisers tested and achieves a final test accuracy only marginally worse than Gadam (but still stronger than all others). On the other hand, in PRN-110, perhaps due to the specific architecture choice, the initial difference in convergence speed of all optimisers is minimal, but Gadam + LH clearly performs very promisingly in the end: it is not only stronger than our result without Lookahead in Figure \ref{fig:lookaheadvgg}(b), but also, by visual inspection, significantly stronger than the SGD + LH results on the same data-set and using the same architecture reported in the original Lookahead paper \cite{zhang2019lookahead}.

Due to the fact that Lookahead is a very recent creation and our constraint on computational resources, we have not been able to fully test Gadam + LH on a wider range of problems. Nonetheless, we believe that the results obtained here are encouraging, and should merit more in-depth investigations in the future works.



\subsection{Effect of Frequency of Averaging}
\label{sec:avgfreq}
\begin{figure}[h]
	\centering
	\begin{subfigure}{0.49\linewidth}
		\includegraphics[width=1\linewidth]{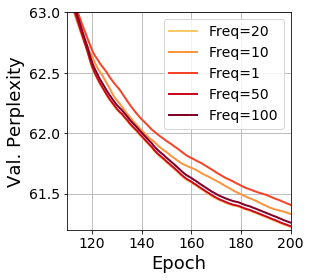}
		\caption{LSTM on PTB}
	\end{subfigure}
	\begin{subfigure}{0.45\linewidth}
		\includegraphics[width=1\linewidth]{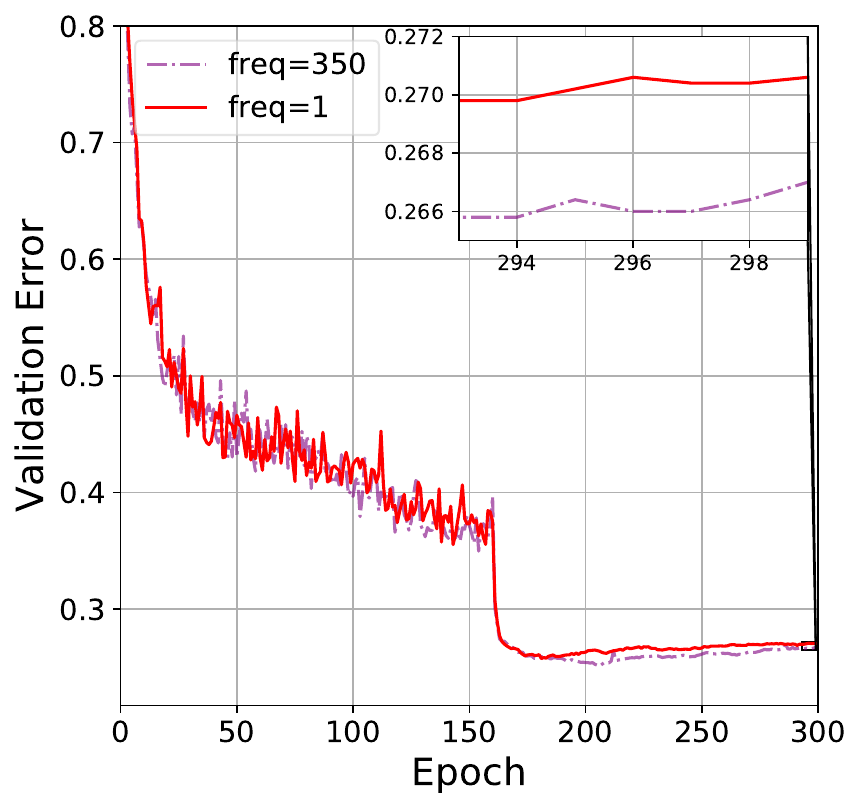}
		\caption{VGG-16 on CIFAR-100}
	\end{subfigure}
	\caption{Effect of different averaging frequencies on validation perplexity of Gadam on representative (a) Language and (b) Image classification tasks. \texttt{Freq=$n$} suggests averaging once per $n$ iterations. \texttt{freq=350} in (b) is equivalently averaging once per \textit{epoch}.} \label{fig:lstm_freq}
\end{figure}

While we derive the theoretical bounds using Polyak-style averaging on every \textit{iteration}, practically we average \textit{much} less: we either average once per \textit{epoch} similar to \cite{izmailov2018averaging}, or select a rather arbitrary value such as averaging once per 100 iterations. The reason is both practical and theoretical: averaging much less leads to significant computational savings, and at the same time as we argued more independent iterates the benefit from averaging is better. In this case, averaging less causes the iterates to be further apart and more independent, and thus fewer number of iterates is required to achieve the similar level of performance if less independent iterates are used. We verify this both on the language and the vision experiments using the identical setup as the main text. With reference to Figure \ref{fig:lstm_freq}(a), not only is the final perplexity very insensitive to averaging frequency (note that the y-axis scale is very small), it is also interesting that averaging \textit{less} actually leads to a slightly better validation perplexity compared to schemes that, say, average every iteration. We see a similar picture emerges in Figure \ref{fig:lstm_freq}(b), where the despite of following very close trajectories, averaging every iteration gives a slightly worse testing performance compared to once an epoch and is also significantly more expensive (with a NVIDIA GeForce RTX 2080 Ti GPU, each epoch of training takes around 10s if we average once per epoch but averaging every iteration takes around 20s).

\subsection{Effect of Average Starting Point and GadamAuto}
\label{sec:gadamauto}
\begin{figure}[h!]
	\centering
	\begin{subfigure}{0.45\linewidth}
		\includegraphics[width=1\linewidth]{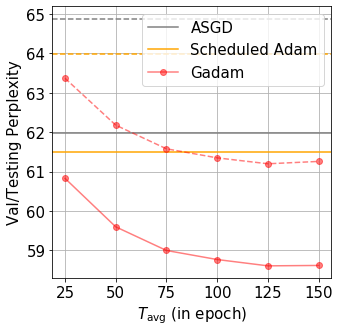}
		\caption{LSTM on PTB. \textit{dashed lines} denote validation and \textit{solid lines} denote test perplexities.}
	\end{subfigure}
	\begin{subfigure}{0.45\linewidth}
		\includegraphics[width=1\linewidth]{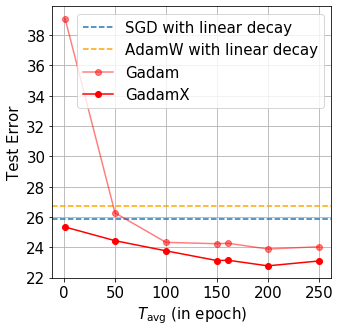}
		\caption{VGG-16 on CIFAR-100}
	\end{subfigure}
	
	\begin{subfigure}{0.45\linewidth}
		\includegraphics[width=1\linewidth]{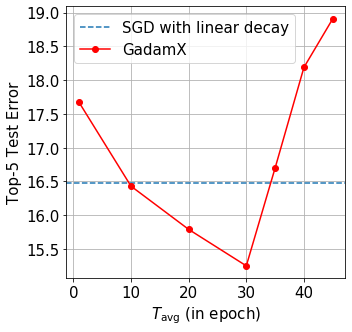}
		\caption{WRN-28-10 on ImageNet 32$\times$32}
	\end{subfigure}
	\begin{subfigure}{0.45\linewidth}
		\includegraphics[width=1\linewidth]{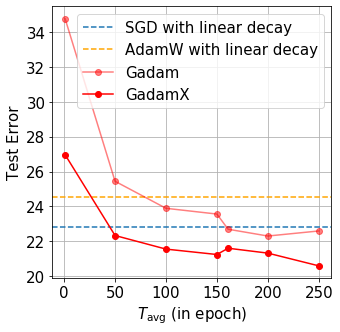}
		\caption{PRN-110 on CIFAR-100}
	\end{subfigure}
	\caption{Effect of different $T_{\mathrm{avg}}$ on the performance of various tasks and architectures}
	 \label{fig:difftavg}
\end{figure}

In Gadam(X), we need to determine when to start averaging ($T_{\mathrm{avg}}$ in Algorithm \ref{alg:Gadam}), and here we investigate the sensitivity of Gadam(X) to this hyperparameter. We use a range of $T_{\mathrm{avg}}$ for a number of different tasks and architectures (Figure \ref{fig:difftavg} and Table \ref{tab:tavg}), including extreme choices such as $T_{\mathrm{avg}} = 0$ (start averaging at the beginning). We observe that for any reasonable $T_{\mathrm{avg}}$, Gadam(X) always outperform their base optimisers with standard learning rate decay, and tuning $T_{\mathrm{avg}}$ yields even more improvements over the heuristics employed in the main text, even if selecting any sensible $T_{\mathrm{avg}}$ already can lead to a promising performance over standard learning rate decay.

\begin{table}[h]
	\caption{Best results obtained from tuning $T_{\mathrm{avg}}$}
	\label{tab:tavg}
	\begin{center}
		\begin{small}
			\begin{sc}
				\begin{tabular}{llc}
					\toprule
					Architecture & Optimiser & Test Acc./Perp.\\
					\midrule
					\multicolumn{3}{l}{\textbf{CIFAR-100}} \\
					VGG-16 & Gadam & 76.11 \\
					& GadamX & 77.22 \\
					PRN-110 & Gadam & 77.41\\
					& GadamX & 79.41 \\
					\midrule
					\multicolumn{3}{l}{\textbf{ImageNet 32$\times$32}} \\
					WRN-28-10 & GadamX &84.75 \\
					\midrule
					\multicolumn{3}{l}{\textbf{PTB}} \\
					LSTM & Gadam & 58.61 \\
					\bottomrule
				\end{tabular}
			\end{sc}
		\end{small}
	\end{center}
	\vskip -0.1in
\end{table}

Here we also conduct preliminary experiments on GadamAuto, a variant of Gadam that uses a constant learning rate schedule and automatically determines the starting point of averaging and training termination - this is possible given the insensitivity of the end-results towards $T_{\mathrm{avg}}$ as shown above, and is desirable as the optimiser both has fewer hyperparameters to tune and trains faster. We use VGG-16 network on CIFAR-100. For all experiments, we simply use a flat learning rate schedule. The results are shown in Table \ref{tab:gadamauto}. We use a patience of 10 for both the determination of the averaging activation and early termination. We also include SWA experiments with SGD iterates.

\begin{table}[h]
	\caption{GadamAuto Test Performance at Termination.}
	\label{tab:gadamauto}
	\begin{center}
		\begin{small}
			\begin{sc}
				\begin{tabular}{llc}
					\toprule
					Optimiser & Data-set & Test Accuracy\\
					\midrule
					Gadam-Auto & CIFAR-100 & 75.39 \\
					SWA-Auto & CIFAR-100 & 73.93 \\
					\bottomrule
				\end{tabular}
			\end{sc}
		\end{small}
	\end{center}
	\vskip -0.1in
\end{table}

It can be seen that while automatic determination for averaging trigger and early termination work well for Gadam (GadamAuto posts a performance only marginally worse than the manually tuned Gadam), they lead to a rather significant deterioration in test in SWA (SWA-Auto performs worse than tuned SWA, and even worse than tuned SGD. This highlights the benefit of using adaptive optimiser as the base optimiser in IA, as the poor performance in SWA-Auto is likely attributed to the fact that SGD is much more hyperparameter-sensitive (to initial learning rate and learning rate schedule, for example. SWA-Auto uses a constant schedule, which is sub-optimal for SGD), and that validation performance often fluctuates more during training for SGD: SWA-Auto determines averaging point based on the number of epochs of validation accuracy stagnation. For a noisy training curve, averaging might be triggered too early; while this can be ameliorated by setting a higher patience, doing so will eventually defeat the purpose of using an automatic trigger. Both issues highlighted here are less serious in adaptive optimisation, which likely leads to the better performance of GadamAuto. 

Nonetheless, the fact that scheduled Gadam still outperforms GadamAuto suggests that there is still ample room of improvement to develop a truly automatic optimiser that performs as strong as or even stronger than tuned ones. One desirable alternative we propose for the future work is the integration of \textit{Rectified Adam} \cite{liu2019variance}, which is shown to be much more insensitive to choice of hyperparameter even compared to Adam.

\section{Experiment Setup}
\label{sec:experimentdetails}
Unless otherwise stated, all experiments are run with PyTorch 1.1 on Python 3.7 Anaconda environment with GPU acceleration. We use one of the three possible GPUs for our experiment: NVIDIA GeForce GTX 1080 Ti, GeForce RTX 2080 Ti or Tesla V100. We always use a single GPU for any single run of experiment.

\subsection{Validating Experiments}
\label{sec:valdetails}

\paragraph{VGG-16 on CIFAR-100}
In this expository experiment, we use the original VGG-16 \textit{without} batch normalisation (batch normalisation has non-trivial impact on conventional measures of sharpness and flatness. See \cite{li2018visualizing}). We conduct all experiments with initial learning rate $0.05$. For fair comparison to previous literature, we use the linear decay schedules advocated in \cite{izmailov2018averaging}, for both SGD and IA. For IA we run the set of terminal learning rates during averaging $\{0.03, 0.01, 0.003\}$, whereas for SGD we decay it linearly to $0.0005$ 

\subsection{Image Classification Experiments} 
\label{sec:imagedetails}
\paragraph{Hyperparameter Tuning} In CIFAR experiments, we tune the base optimisers (i.e. SGD, Adam(W), Padam(W)) only, and assuming that the ideal hyperparameters in base optimisers apply to IA, and apply the same hyperparameter setting for the corresponding IA optimisers (i.e. SWA, Gadam, GadamX). For SGD, we use a base learning rate of 0.1 and use a grid searched initial learning rates in the range of  $\{0.001, 0.01, 0.1\}$ and use the same learning rate for Padam, similar to the procedues suggested in \cite{chen2018closing}. For Adam(W), we simply use the default initial learning rate of $0.001$ except in VGG-16, where we use initial learning rate of $0.0005$. After the best learning rate has been identified, we conduct a further search on the weight decay, which we find often leads to a trade-off between the convergence speed and final performance; again we search on the base optimisers only and use the same value for the IA optimisers. For CIFAR experiments, we search in the range of $[10^{-4}, 10^{-3}]$, from the suggestions of \cite{loshchilov2018decoupled}. For decoupled weight decay, we search the same range for the weight decay scaled by initial learning rate. 

On ImageNet~\citep{russakovsky2015imagenet} experiments, we conduct the following process. On WRN we use the settings recommended by \cite{chrabaszcz2017downsampled}, who conducted a thorough hyperparameter search: we set the learning rate at $0.03$ and weight decay at $0.0001$ for SGD/SWA and Padam, based on their searched optimal values. for AdamW/Gadam, we set decoupled weight decay at $0.01$ and initial learning rate to be $0.001$ (default Adam learning rate). For GadamX, we again use the same learning rate of $0.03$, but since the weight decay in GadamX is partially decoupled, we set the decoupled weight decay to $0.0003$. On PRN-110, we follow the recommendations of the authors of \cite{he2016identity} to set the initial learning rate for SGD, Padam and GadamX to be $0.1$. For AdamW and Gadam, we again use the default learning rate of $0.001$. Following the observation by \cite{loshchilov2018decoupled} that smaller weight decay should be used for longer training (in PRN-110 we train for 200 epochs), we set weight decay at $10^{-5}$ and decoupled weight decay at $0.0003$ (GadamX)/$0.001$ (others) respectively, where applicable.

Overall, we do \textbf{not} tune adaptive methods (Adam and Gadam) as much (most noticeably, we usually fix their learning rate to 0.001), and therefore in particular the AdamW results we obtain may or may not be at their optimal performance. Nonetheless, the rationale is that by design, one of the key advantage claimed is that adaptive optimiser should be less sensitive to hyperparameter choice, and in this paper, the key message is that Gadam performs well, \textit{despite of AdamW, its base optimiser, is rather crudely tuned}.

In all experiments, momentum parameter ($\beta = 0.9$) for SGD and $\{\beta_1, \beta_2\} = \{0.9, 0.999\}$ and $\epsilon = 10^{-8}$ for Adam and its variants are left at their respective default values. For all experiments unless otherwise stated, we average once per epoch. We also apply standard data augmentation (e.g. flip, random crops) and use a batch size of 128 for all experiments conducted.

\begin{table*}[t]
	\caption{Baseline Results from Previous Works}
	\label{tab:baselines}
	\begin{center}
		\begin{small}
			
			\begin{sc}
				\begin{tabular}{llccc}
					\toprule
					
					Network & Optimiser & Accuracy/Perplexity & Reference \\
					\midrule
					\textbf{CIFAR-100}\\
					VGG-16 & SGD & 73.80 & \cite{huang2018data} \\
					VGG-16 & FGE & 74.26 & \cite{izmailov2018averaging}\\
					PRN-164 & SGD & 75.67 & \cite{he2016identity} \\
					PRN-110 & SGD & 76.35 & online repository** \\
					ResNet-164 & FGE & 79.84 & \cite{izmailov2018averaging} \\
					ResNeXt-29 & SGD & 82.20 & \cite{xie2017aggregated} \\
					ResNeXt-29 & SGD & 81.47 & \cite{bansal2018can} \\
					\midrule
					\textbf{CIFAR-10}\\
					VGG-19 & SGD & 93.34 & online repository** \\
					VGG-16 & SGD & 93.90 & \cite{huang2018data} \\
					PRN-110 & SGD & 93.63 & \cite{he2016identity} \\
					PRN-110 & SGD & 95.06 & online repository** \\
					\midrule
					\textbf{ImageNet 32$\times$32}\\
					WRN-28-10 & SGD & 59.04/81.13* &
					\cite{chrabaszcz2017downsampled} \\
					Modified WRN & SGD & 60.04/82.11* & \cite{mcdonnell2018training} \\
					\midrule
					\textbf{PTB} \\
					LSTM 3-layer & NT-ASGD & 61.2/58.8*** & \cite{merity2017regularizing} \\
					\bottomrule
					\multicolumn{4}{l}{\textbf{Notes:}
					}\\
					\multicolumn{4}{l}{* Top-1/Top-5 Accuracy
					}\\
					\multicolumn{4}{l}{** Link: \url{https://github.com/bearpaw/pytorch-classification}
					}\\
					\multicolumn{4}{l}{*** Validation/Test Perplexity
					}\\
				\end{tabular}
			\end{sc}
		\end{small}
	\end{center}
	\vskip -0.1in
\end{table*}

\paragraph{Learning Rate Schedule} For all experiments without IA,  we use the following learning rate schedule for the learning rate at the $t$-th epoch, similar to \cite{izmailov2018averaging}, which we find to perform better than the conventionally employed step scheduling (refer to the experimental details in Appendix \ref{sec:linearvsstep}):
\begin{equation}
	\alpha_t = 
	\begin{cases}
		\alpha_0, & \text{if}\ \frac{t}{T} \leq 0.5 \\
		\alpha_0[1 - \frac{(1 - r)(\frac{t}{T} - 0.5)}{0.4}] & \text{if } 0.5 < \frac{t}{T} \leq 0.9 \\
		\alpha_0r, & \text{otherwise}
	\end{cases}
\end{equation}
where $\alpha_0$ is the initial learning rate. In the motivating logistic regression experiments on MNIST, we used $T = 50$. $T = 300$ is the total number of epochs budgeted for all CIFAR experiments, whereas we used $T = 200$ and $50$ respectively for PRN-110 and WideResNet 28x10 in ImageNet. We set $r = 0.01$ for all experiments. For experiments with iterate averaging, we use the following learning rate schedule instead:
\begin{equation}
	\alpha_t = 
	\begin{cases}
		\alpha_0, & \text{if}\ \frac{t}{T_{\mathrm{avg}}} \leq 0.5 \\
		\alpha_0[1 - \frac{(1 - \frac{\alpha_{\mathrm{avg}}}{\alpha_0})(\frac{t}{T} - 0.5)}{0.4}] & \text{if } 0.5 < \frac{t}{T_{\mathrm{avg}}} \leq 0.9 \\
		\alpha_{\mathrm{avg}}, & \text{otherwise}
	\end{cases}
\end{equation}
where $\alpha_{\mathrm{avg}}$ refers to the (constant) learning rate after iterate averaging activation, and in this paper we set $\alpha_{\mathrm{avg}} = \frac{1}{2}\alpha_0$. $T_{\mathrm{avg}}$ is the epoch after which iterate averaging is activated, and the methods to determine $T_{\mathrm{avg}}$ was described in the main text. This schedule allows us to adjust learning rate smoothly in the epochs leading up to iterate averaging activation through a similar linear decay mechanism in the experiments without iterate averaging, as described above.

The only exception is the WRN experiments on ImageNet 32$\times$32, where we only run 50 epochs of training and start averaging from 30th epoch. We found that when using the schedule described above for the IA schedules (SWA/Gadam/GadamX), we start decay the learning rate too early and the final result is not satisfactory. Therefore, for this particular set of experiments, we use the same learning rate schedule for both averaged and normal optimisers. The only difference is that for IA experiments, we decay the learning rate until the 30th epoch and keep it fixed for the rest of the training.

\subsection{Language Modelling Experiments}
\label{sec:languagedetails}
In language modelling experiments, we use the codebase provided by \url{https://github.com/salesforce/awd-lstm-lm}. For ASGD, we use the hyperparameters recommended by \cite{merity2017regularizing} and set the initial learning rate to be 30. Note that in language experiments, consistent with other findings decoupled weight decay seems to be not as effective $L_2$, possibly due to LSTM could be more well-regularised already, and that batch normalisation, which we argue to be central to the efficacy of decoupled weight decay, is not used in LSTM. Thus, for this set of experiments we simply use Adam and Padam as the iterates for Gadam and GadamX. For Adam/Gadam, we tune the learning rate by searching initial learning rate in the range of $\{0.0003, 0.001, 0.003, 0.01\}$ and for Padam and GadamX, we set the initial learning rate to be $1$ and partially adaptive parameter $p = 0.2$, as recommended by the authors \cite{chen2018closing}. We further set the weight decay to be their recommended value of $1.2 \times 10^{-6}$. For the learning rate schedule, we again follow \cite{merity2017regularizing} for a piecewise constant schedule, and decay the learning rate by a factor of $10$ at the $\{100, 150\}$-th epochs for all experiments without using iterate averaging. For experiments with iterate averaging, instead of decaying the learning rate by half before averaging starts, we keep the learning rate constant thoughout to make our experiment comparable with the ASGD schedule. We run all experiments for $200$ (instead of $500$ in \cite{merity2017regularizing}) epochs.

\paragraph{Learning Rate Schedule} As discussed in the main text, the experiments shown in Table \ref{tab:language} and Figure \ref{fig:pennvalperplexity} are run with constant schedules (except for Padam). Padam runs with a step decay of factor of 10 at $\{100, 150\}$-th epochs. However, often even the adaptive methods such as Adam are scheduled with learning rate decay for enhanced performance. Therefore, we also conduct additional scheduled experiments with Adam, where we follow the same schedule of Padam. The results are shown in Appendix \ref{sec:lstmschedule}.

\subsection{Experiment Baselines} 
\label{sec:baselines}
To validate the results we obtain and to make sure that any baseline algorithms we use are properly and fairly tuned, we also survey the previous literature for baseline results where the authors use same (or similar) network architectures on the same image classification/language tasks, and the comparison of our results against theirs is presented in Table \ref{tab:baselines}. It is clear that for most of the settings, our baseline results achieve similar or better performance compared to the previous work for comparable methods; this validates the rigour of our tuning process.  
\section{Lessons learned from ImageNet}
\begin{figure}[h!]
	\begin{subfigure}[b]{0.49\linewidth}
		\includegraphics[width=\textwidth]{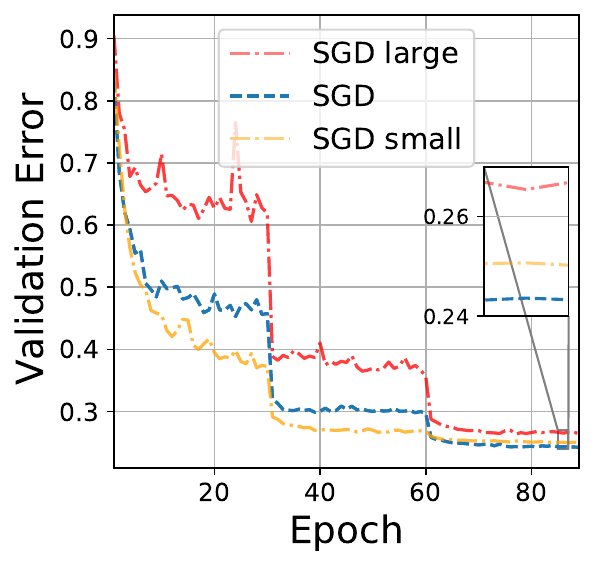}
		\caption{SGD $\alpha=[0.03,0.1,0.3]$}
		\label{subfig:sgdneedsideal}
	\end{subfigure}
	\begin{subfigure}[b]{0.49\linewidth}
		\includegraphics[width=\textwidth]{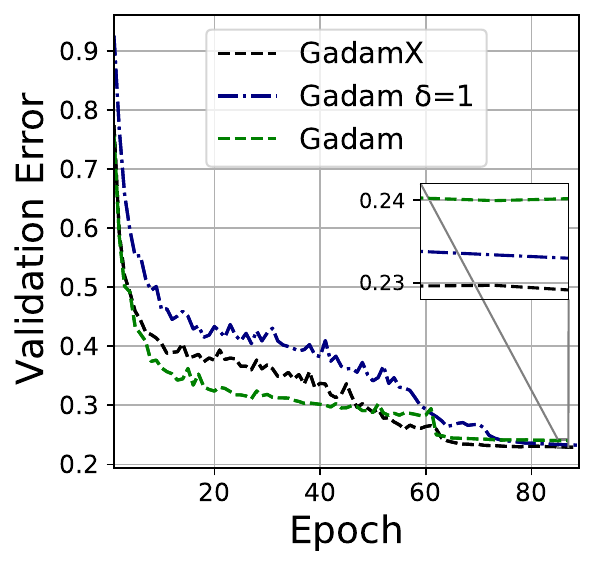}
		\caption{Speed/Error Trade-off}
		\label{subfig:nofreelunch}
	\end{subfigure}
	\caption{(a) Unlike IA adaptive methods, SGD does not benefit from larger initial learning rates. (b) To attain the greatest generalisation with adaptive methods, the fast convergence is sacrificed. Gadam $\delta=1$, has a correspondingly large learning rate $0.5$}
	\label{fig:nofreelunch}    
	\vspace{-10pt}
\end{figure}
\paragraph{1. Adaptive IA makes use of huge initial learning rates:}
unlike SGD and SWA, which have a strong performance degradation when large initial learning rates are used, shown in Fig~\ref{subfig:sgdneedsideal}, we find that large initial learning rates improve the generalisation performance of Gadam/GadamX, with the largest initial learning rates giving the best results. 
\paragraph{2. Convergence speed comes at a cost:} Combining a large numerical stability coefficient and large learning rates allows Gadam to give significantly superior performance to SGD. However, the price paid is in convergence speed, shown in Fig~\ref{subfig:nofreelunch}. For these settings the convergence speed is often as slow or slower than GadamX. Using the same settings as in the small scale experiments (shown as Gadam in the graph) we achieve a top-$1$ accuracy of $75.52$ for ResNet-$50$. Whilst this significantly improves upon the base optimiser AdamW, these results are not as strong as those of SGD. Whilst increasing the base learning rate to $0.003$ increases the ResNet-$50$ Gadam generalisation performance to $76.53$, much of the convergence speed is already lost. We note that the effective weight decay is given by $(1-\alpha\gamma)$ so we expect higher regularisation from higher learning rates. We do not find that increasing the weight decay whilst keeping the same base learning rates produces as strong results in our experiments and hence this learning rate and weight decay interplay could form the basis for interesting future work.
\paragraph{3. Partially adaptive optimisation generalises best:}
We find that for all experiments GadamX delivers the strongest performance. We do not find a strong dependence on the choice of the IA starting point (we try epoch $61,71,81$). We find that altering the numerical stability constant gives a small boost in Top-$1$ error, from $77.19$ to $77.31$ for the ResNet-$50$, but that results remain strong for the traditional setting.
\paragraph{Comparison to previous results:} We specifically report the final (as opposed to best) validation error for all our runs. We find the best SGD ResNet-$50$/$101$ results to be $75.75/77.62\%$, which are slightly worse/better than the official repository results. All of these results are still significantly lower than results achieved by Gadam/GadamX. We note that iterate averaged methods seem to continually decrease error in the final epochs of training, unlike SGD, which can sometimes overfit slightly in the final epochs of training.
\section{Additional Experimental Results}
\label{sec:additionalexperiments}
\subsection{ResNext CIFAR-$100$ and WideResNet$28\times10$ on 32$\times$32 ImageNet}
\begin{table}[!htb]
\begin{minipage}{.45\linewidth}
\caption{Test Accuracy on CIFAR-$100$.}
\begin{center}
   \begin{small}
	\begin{scriptsize}
	\begin{tabular}{llcc}
		\toprule
		Architecture & Optimiser & Test Accuracy \\
		\midrule
		ResNeXt-29 & SGD & 81.47$\pm 0.17$ & \\
		& SWA & 82.95$\pm 0.28$ & \\
		& Adam(W) &
		80.16$\pm 0.16$& \\
		& Padam(W) &
		82.37$\pm 0.35$ & \\
		& Gadam & 82.13$\pm 0.20$ &\\
		& GadamX & \textbf{83.27}$\pm 0.11$ &\\
		\bottomrule
	\end{tabular}
	\end{scriptsize}
\end{small}
\end{center}
\hfill
\end{minipage}
\begin{minipage}{.55\linewidth}
\caption{Test Accuracy on ImageNet 32$\times$32.}
	\label{tab:imagenet}
	\begin{center}
		\begin{small}
			\begin{scriptsize}
				\begin{tabular}{llcc}
					\toprule
					Architecture & Optimiser & Top-1 & Top-5 \\
					\midrule
					WRN-28-10 & SGD & 61.33$\pm 0.11$& 83.52$\pm0.14$  \\
					& SWA & 62.32$\pm 0.13$& 84.23$\pm0.05$ \\
					& AdamW & 55.51$\pm0.19$ & 79.09$\pm0.33$ \\
					& Padam & 59.65$\pm 0.17$ & 81.74$\pm0.16$ \\
					& Gadam & 60.50$\pm 0.19$ & 82.56$\pm0.13$\\
					& GadamX & \textbf{63.04}$\pm0.06$ & \textbf{84.75}$\pm0.03$\\
					\bottomrule
				\end{tabular}
			\end{scriptsize}
		\end{small}
	\end{center}
	\vskip -0.1in
\end{minipage}
\end{table}

\subsection{Testing Performance of CIFAR-10}
\label{sec:c10}
\begin{figure}[h]
	\centering
	\begin{subfigure}[b]{0.23\textwidth}
		\includegraphics[width=\textwidth]{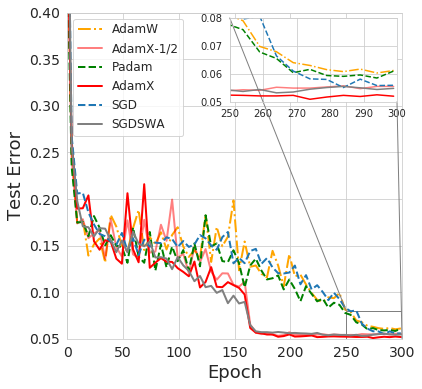}
		\caption{VGG-16}
	\end{subfigure}
	\begin{subfigure}[b]{0.23\textwidth}
		\includegraphics[width=\textwidth]{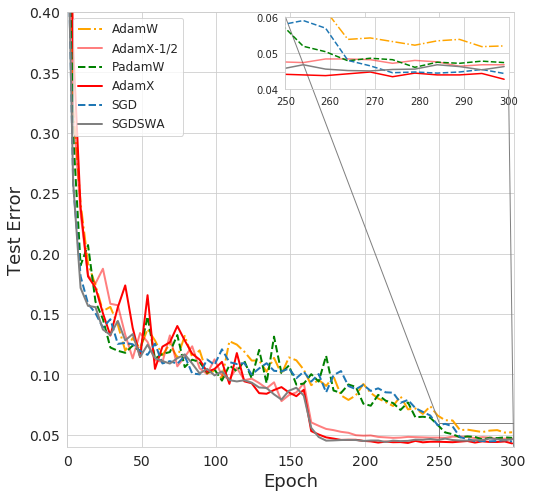}
		\caption{PRN-110}
	\end{subfigure}
	\caption{Test Error on CIFAR-10}\label{fig:c10}
\end{figure}

We report the testing performance of VGG-16 and PRN-110 on CIFAR-10 in Figure \ref{fig:c10} and Table \ref{tab:c10}. Perhaps due to the fact that CIFAR-10 poses a simpler problem compared to CIFAR-100 and ImageNet in the main text, the convergence speeds of the optimisers differ rather minimally. Nonetheless, we find that GadamX still outperforms all other optimisers by a non-trivial margin in terms of final test accuracy.

\begin{table}[h]
	\caption{Top-1 Test Accuracy on CIFAR-10 Data-set}
	\label{tab:c10}
	\begin{center}
		\begin{small}
			\begin{sc}
				\begin{tabular}{llcc}
					\toprule
					Architecture & optimiser & Test Accuracy \\
					\midrule
					VGG-16    & SGD & 94.14$\pm0.37$ &  \\
					& SWA & 94.69$\pm0.36$ & \\
					& Adam(W) & 93.90 $\pm0.11$ & \\
					& Padam(W) & 94.13 $\pm0.06$ & \\
					& Gadam & 94.62$\pm0.15$ &\\
					& GadamX & \textbf{94.88}$\pm0.03$&\\
					\midrule
					PRN-110 & SGD & 95.40$\pm0.25$& \\
					& SWA & 95.55$\pm0.12$& \\
					& Adam(W) & 94.69$\pm0.14$ & \\
					& Padam(W) & 95.28$\pm0.13$ & \\
					& Gadam & 95.27$\pm0.02$ &\\
					& GadamX & \textbf{95.95$\pm0.06$} &\\
					\bottomrule
				\end{tabular}
			\end{sc}
		\end{small}
	\end{center}
	\vskip -0.1in
\end{table}

\subsection{Word Level Language Modelling with Learning Rate Schedules and Non-monotonic Trigger}
\label{sec:lstmschedule}

\paragraph{Word-level Language Modelling on PTB}
{\begin{figure}[h!]
		\centering
		\includegraphics[width=0.8\linewidth]{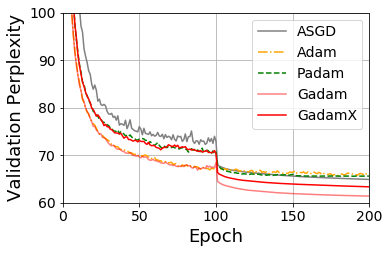}
		\caption{Validation perplexity of 3-layer LSTM on PTB word-level modelling }
		\label{fig:pennvalperplexity}
\end{figure}}
We run word-level language modelling using a 3-layer Long-short Term Memory (LSTM) model \cite{gers1999learning} on PTB dataset \cite{marcus1993building} and the results are in Table \ref{tab:language} and Fig. \ref{fig:pennvalperplexity}. Remarkably, Gadam achieves a test perplexity of 58.77 (58.61 if we tune $T_{\mathrm{avg}}$. See Table \ref{tab:tavg} in App. \ref{sec:gadamauto}), better than the baseline NT-ASGD in \cite{merity2017regularizing} that \textit{runs an additional 300 epochs} on an identical network. Note that since, by default, the ASGD uses a constant learning rate, we do \textit{not} schedule the learning rate except Padam which requires scheduling to converge. Also, for consistency, we use a manual trigger to start averaging at the 100th epoch for ASGD (which actually outperforms the NT-ASGD variant). We additionally conduct experiments \textit{with} scheduling and NT-ASGD (App. \ref{sec:additionalexperiments}) and Gadam still outperforms. It is worth mentioning that for state of the art results in language modelling \cite{melis2017state,brown2020language,shoeybi2019megatron}, Adam is the typical optimiser of choice. Hence these results are both encouraging and significant for wider use in the community.

\begin{figure}[h]
	\centering
	\includegraphics[width=0.8\linewidth]{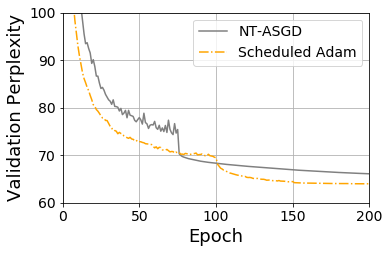}
	\vspace*{-5mm}
	\caption{Validation Perplexity of NT-ASGD and Scheduled Adam on 3-layer LSTM PTB Word-level Modelling.}    \label{fig:pennvalperplexity2}
\end{figure}

\begin{table}[t]
	\caption{Validation and Test Perplexity on Word-level Language Modelling.}
	\label{tab:language}
	\begin{center}
		\begin{small}
			\begin{sc}
				\begin{tabular}{llcc}
					\toprule
					
					Data-set & optimiser &\multicolumn{2}{c}{Perplexity} \\
					& & Validation & Test \\
					\midrule
					PTB & ASGD & 64.88$\pm0.07$ & 61.98$\pm0.19$ \\
					& Adam & 65.96$\pm0.08$ & 63.16$\pm0.24$ \\
					& Padam & 65.69$\pm0.07$& 62.15$\pm0.12$ \\
					& Gadam & \textbf{61.35}$\pm0.05$& \textbf{58.77}$\pm0.08$\\
					& GadamX & 63.49$\pm0.19$ & 60.45$\pm0.04$\\
					\bottomrule
				\end{tabular}
			\end{sc}
		\end{small}
	\end{center}
	\vskip -0.1in
\end{table}
Here we include additional results on word-level language modelling using \textit{scheduled} Adam and NT-ASGD, where the point to start averaging is learned non-monotonically and automatically. Where scheduling further improves the Adam performance marginally, the automatically triggered ASGD actually does not perform as well as the manually triggered ASGD that starts averaging from 100th epoch onwards, as we discussed in the main text - this could be because that ASGD converges rather slowly, the 200-epoch budget is not sufficient, or the patience (we use patience = 10) requires further tuning. Otherwise, our proposed Gadam and GadamX without IA schedules still outperform the variants tested here \textit{without careful learning rate scheduling}. The results are summarised in Figure \ref{fig:pennvalperplexity2} and Table \ref{tab:language2}.

\begin{table}[t]
	\caption{Validation and Test Perplexity on Word-level Language Modelling. The Gadam(X) results are lifted from Table \ref{tab:language}.}
	\label{tab:language2}
	\begin{center}
		\begin{small}
			\begin{sc}
				\begin{tabular}{llcc}
					\toprule
					
					Data-set & optimiser &\multicolumn{2}{c}{Perplexity} \\
					& & Validation & Test \\
					\midrule
					PTB & NT-ASGD & 66.01 &  64.73\\
					& Scheduled Adam & 63.99 & 61.51 \\
					\midrule
					& Gadam (Ours) & \textbf{61.35}& \textbf{58.77}\\
					& GadamX (Ours) & 63.49 & 60.45\\
					\bottomrule
				\end{tabular}
			\end{sc}
		\end{small}
	\end{center}
	\vskip -0.1in
\end{table}

\subsection{Relation between
	Improvement from Averaging and Number of Parameters in Previous Work}
\label{sec:iaprevious}

\begin{figure}[h]
	\centering
	\includegraphics[width=0.7\linewidth]{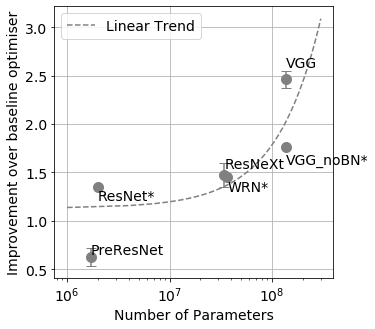}
	\vspace*{-5mm}
	\caption{Number of parameters $P$ against improvement margin for both results obtained by us and in \cite{izmailov2018averaging} (annotated with asterisks) on CIFAR-100}    \label{fig:iaprevious}
\end{figure}

In this section we demonstrate that our claim that there should be a dependence on number of parameters $P$ on the margin of improvement from averaging is also present in previous works that use IA or a related ensemble method. Here we use the results from Table 1 of \cite{izmailov2018averaging}. Since the different network architectures are trained with different budget of epochs which make the direct comparison of SWA results difficult, we instead consider their FGE \cite{garipov2018loss} results which the author argue to have the similar properties to and that is actually approximated by SWA. We plot their result along with us in Figure \ref{fig:iaprevious}. While we do not obtain a perfect linear relationship possibly due to a wide range of possible interfering factors such as difference in architecture, use of batch normalisation, choice of optimiser and hyperparameter tuning, again we nevertheless observe that there exists a roughly positive relationship between $P$ and the margin of test improvement.

\subsection{Linear vs Step Scheduling}
\label{sec:linearvsstep}
In this work, for the \textit{baseline} methods in image classification tasks we use \textit{linear} instead of the more conventionally employed \textit{step} scheduling because we find linear scheduling to generally perform better in the experiments we conduct. In this section, we detail the results of these experiments, and in this section, `linear' refers to the schedule introduced in Appendix \ref{sec:imagedetails} and `step' refers to the schedule that reduces the learning rate by a factor of 10 in $\{150, 250\}$ epochs for 300-epoch experiments (CIFAR datasets), or in $\{25, 40\}$ epochs for 50-epoch experiments (ImageNet dataset). The results are shown in Table \ref{tab:linearvsstep}.

\begin{table}[h]
	\caption{Testing performance of linear and step learning rate schedules on baseline methods.}
	\label{tab:linearvsstep}
	\begin{center}
		\begin{small}
			\begin{sc}
				\begin{tabular}{llcc}
					\toprule
					Architecture & Optimiser & Step & Linear \\
					\midrule
					\multicolumn{4}{l}{\textbf{CIFAR-100}} \\
					VGG-16 & SGD & 73.28 & \textbf{74.15} \\
					& AdamW &  73.20  & \textbf{73.26}\\
					& Padam & 74.46 & \textbf{74.56}\\
					PRN-110 & SGD & \textbf{77.23} & 77.22 \\
					& AdamW & 75.27 & \textbf{75.47}\\
					& Padam & 73.95 & \textbf{77.30}\\
					\bottomrule
				\end{tabular}
			\end{sc}
		\end{small}
	\end{center}
	\vskip -0.1in
\end{table}

\newpage
\section{Importance of Iterate Averaging for Convergence}
\label{sec:optimisationaveraging}
We argue that despite of the universal practical use of the final iterate of optimisation, it is heuristically motivated and in most proofs of convergence, some form of iterative averaging is required and used implicitly to derive the theoretical bounds. For $\beta$-Lipschitz, convex empirical risks, denoted the (overall) loss $L$. The difference between the $t+1$'th iterate and the optimal solution $L_{\vw}^{*}$ can be bounded. The sum of differences along the trajectory (known as the \textit{regret}) telescopes, hence resulting in a convergence rate for the average regret which is an upper bound for the loss of the average point \cite{nesterov2013introductory,duchi2018introductory}: 
\begin{equation}
	\begin{aligned}
		& \delta L =  L_{\vw_{t+1}
		}-L_{\vw^{*}} \leq \nabla L_{\vw_{t}}(\vw_{t+1}-\vw^{*}) +  \frac{\beta}{2}||\vw_{t+1}-\vw_{t}||^{2}\\
		&  \mathbb{E}(\delta L) \leq \hat{\nabla} L _{\vw_{t}}(\vw_{t}-\vw^{*})-(\alpha-\frac{\beta \alpha^{2}}{2})||\hat{\nabla} L_{\vw_{t}} ||^{2} + \alpha \sigma^{2}_{t}   \\
	\end{aligned}
\end{equation}
where $\hat{\nabla}L_{\vw_{t}}$ is the noisy gradient at $\vw_{t}$ and $\sigma^{2}_{t}$ is its variance: $\text{Var}( \hat{\nabla}L_{\vw_{t}})$. Noting that $\vw_{t+1} = \vw_{t} - \alpha\hat{\nabla} L_{\vw_{t}}$:
\begin{equation}
	\begin{aligned}
		\label{eq:proofsketch}
		\frac{R}{T} & = \mathbb{E}\Big[\frac{1}{T}\sum_{t=1}^{T-1}L_{\vw_{t+1}}-L_{\vw^{*}}\Big] \\
	\end{aligned}
\end{equation}
Using Jensen's inequality, we have:
\begin{equation}
	\begin{aligned}
		\frac{R}{T} &  \leq \frac{1}{T}\sum_{t=0}^{T-1}\frac{||\vw_{t}-\vw^{*}||^{2}-||\vw_{t+1}-\vw^{*}||^{2}}{2\alpha} + \alpha \sigma^{2}_{t} \\
		& \mathbb{E}[L_{\frac{1}{T}\sum_{t=1}^{T-1}\vw_{t+1}}-L_{\vw^{*}}]\leq \frac{R}{T} \leq \frac{||\vw_{0}-\vw^{*}||^{2}}{2\alpha T}+\alpha\sigma^{2}_{m} \\
	\end{aligned} 
\end{equation}
where $\sigma^{2}_{m} = \argmax_{\vw_{t}} \mathbb{E}||\hat{\nabla}L_{\vw_{t}}-\nabla L_{\vw_{t}}||^{2}$, and $R$ is the regret. Setting $\alpha = (\beta+\sigma\frac{\sqrt{T}}{D})^{-1}$ in \eqref{eq:proofsketch} gives us the optimal convergence rate. Similar convergence results can be given for a decreasing step size $\alpha_{t} \propto t^{-1/2}\alpha_{0}$. For adaptive optimisers, the noisy gradient is preconditioned by some non-identity matrix $\bar\mB^{-1}$:
\begin{equation}
	\vw_{k+1} \leftarrow \vw_{k} - \alpha \bar\mB^{-1} \nabla L_k(\vw)
\end{equation}
Methods of proof \cite{reddi2019convergence, tran2019convergence} rely on bounding the regret $\mathcal{O}(\sqrt{T})$ and showing that the average regret $\frac{R}{T} \rightarrow 0$ and Equation \ref{eq:proofsketch} explicitly demonstrates that the average regret is an upper bound on the expected loss for the average point in the trajectory. Hence existing convergence results in the literature prove convergence for the iterate average, but not the final iterate.
\subsubsection{Optimal Learning Rates}
Setting $\alpha = (\beta+\sigma\frac{\sqrt{T}}{D})^{-1}$ gives us the optimal convergence rate of $\frac{\beta R^{2}}{T}+\frac{\sigma D}{\sqrt{T}}$. Similar convergence results can be given for a decreasing step size $\alpha_{t} \propto t^{-1/2}\alpha_{0}$ \cite{duchi2018introductory} when the number of iterations $T$ is not known in advance. Given the use of both iterate averaging and learning rate schedule in the proofs, it is difficult to understand the relative importance of the two and how this compares with the typical heuristic of using the final point.

\subsection{Relevance of Local Geometry Arguments}
\label{sec:geometry}
One argument as to why IA improves generalisation \cite{izmailov2018averaging}  is about the local geometry of the solution found: \cite{izmailov2018averaging} discuss the better generalisation of SWA to the ``flatter'' minimum it finds. The same argument is used to explain the apparent worse generalisation of adaptive method: \cite{wu2018sgd} showed empirically that adaptive methods are not drawn to flat minima unlike SGD. From both Bayesian and minimum description length arguments \cite{hochreiter1997flat}, flatter minima generalise better, as they capture more probability mass. \cite{he2019asymmetric} formalise the intuition under the assumption of a shift between the training and testing loss surface and investigate the presence of ``flat valleys'' in loss landscape. They argue that averaging leads to a biased solution to the ``flatter'' valley, which has \textit{worse} training but \textit{better} generalisation performance due to the shift. This suggests IA has an inherent regularising effect, which contrasts with our previous claim that IA should improve both. 

However, one issue in the aforementioned analysis, is that they train their SGD baseline and averaged schemes on different learning rate schedules. While this is practically justified, and even desirable, exactly because IA performs better with high learning rate as argued, for \textit{theoretical analysis} on the relevance of the landscape geometry to solution quality, it introduces interfering factors. It is known that the learning rate schedule can have a significant impact on both performance and curvature \cite{jastrzebski2020the}. We address this by considering IA and the iterates, for the same learning rate to specifically alleviate this issue. We use the VGG-16 \emph{without} BN\footnote{It is argued that BN impacts the validity of conventional measures of sharpness \cite{liu2019variance} hence we deliberately remove BN here, nor do we tune optimisers rigorously, since the point here is for theoretical exposition instead of empirical performance. See detailed setup in App. \ref{sec:imagedetails}.} using both AdamW/Gadam and SGD/SWA. In addition to the test and training statistics, we also examine the spectral norm, Frobenius norm and trace which serve as different measures on the ``sharpness'' of the solutions using the spectral tool by \cite{granziol2019mlrg}; we show the results in Table \ref{table:vggsharp}.
\begin{table*}[h!]
	\centering
	\caption{Performance and Hessian-based sharpness metrics on CIFAR-100 using VGG-16. The numerical results for iterates are in brackets.}
	\begin{scriptsize}
		\begin{tabular}{@{}ccccccc@{}}
			\toprule
			Optimiser & Terminal LR & Train acc. & Test acc. & Spectral Norm & Frobenius Norm & Trace \\ 
			\midrule
			AdamW & $3E{-6}$ & 99.93 & 69.43 & 62 & $9.3E{-4}$ & $4.7E{-5}$ \\
			Gadam & $3E{-5}$ & \textbf{99.97} (94.12) & 69.67 (67.16) & 120 (2500) & $1.4E{-3}$(0.86) & $6.4E{-5}$($2.2E{-3}$) \\
			Gadam &$3E{-4}$ & 98.62 (89.34) & \textbf{71.55} (64.68) & 43 (280) & $1.1E{-3}$ (0.023) &$1.1E{-4}$ ($5.1E{-4}$)\\
			\midrule
			SGD & $3E{-4}$ & 99.75 & 71.64 &  4.40 & $1.2E{-5}$ & $4.7E{-6}$\\
			SWA &$3E{-3}$ & \textbf{99.98} (98.87) & 71.32 (69.88) & 1.85 (14.6) & $4.4E{-6}$ ($1.3E{-4}$) & $1.1E{-6}$ ($8.6E{-5}$)\\
			SWA & $3E{-2}$ & 91.58 (77.29) & \textbf{73.40} (63.42) & 1.35 (12.0) & $8.4 E{-6}$ ($7.0E{-5}$)& $1.8E{-5}$ ($9.8E{-5}$)\\
			\bottomrule
			
		\end{tabular}
		\label{table:vggsharp}
	\end{scriptsize}
\end{table*}
We find a rather mixed result with respect to the local geometry argument. While averaging indeed leads to solutions with lower curvature, we find no clear correlation between flatness and generalisation. One example is that compared to SGD, the best performing Gadam run has $14\times$ larger spectral norm, $92\times$ larger Frobenius norm and $23\times$ larger Hessian trace, yet the test accuracy is only $0.09\%$ worse. Either our metrics do not sufficiently represent sharpness, which is unlikely since we included multiple metrics commonly used, or that it is not the most relevant \textit{explanation} for the generalisation gain. We hypothesise the reason here is that the critical assumption, upon which the geometry argument builds, that there exist only \textit{shifts} between test and train surfaces is unsound despite a sound analysis \textit{given} that. For example, recent work has shown under certain assumptions that the true risk surface is \textit{everywhere} flatter than the empirical counterpart \cite{granziol2020towards}. Furthermore, for any arbitrary learning rate, as predicted IA helps \textit{both} optimisation and generalisation \textit{compared to iterates of the same learning rate}; any trade-offs between optimisation and generalisation seem to stem from the choice of \textit{learning rates} only.

\end{document}

%% file: math.tex
\usepackage{amsmath,amsfonts,bm}

\def\eqref#1{equation~\ref{#1}}

\def\1{\bm{1}}

\def\vphi{{\bm{\phi}}}
\def\vepsilon{{\bm{\epsilon}}}

\def\vg{{\bm{g}}}

\def\vm{{\bm{m}}}

\def\vv{{\bm{v}}}
\def\vw{{\bm{w}}}
\def\vx{{\bm{x}}}

\def\mB{{\bm{B}}}

\def\mH{{\bm{H}}}

\def\mX{{\bm{X}}}

\def\mLambda{{\bm{\Lambda}}}

\DeclareMathAlphabet{\mathsfit}{\encodingdefault}{\sfdefault}{m}{sl}
\SetMathAlphabet{\mathsfit}{bold}{\encodingdefault}{\sfdefault}{bx}{n}

\DeclareMathOperator*{\argmax}{arg\,max}

\DeclareMathOperator{\Tr}{Tr}

%% file: neurips_2021_checklist.tex
\section*{Checklist}

\begin{enumerate}

\item For all authors...
\begin{enumerate}
  \item Do the main claims made in the abstract and introduction accurately reflect the paper's contributions and scope? \\
    \answerYes{} \textit{Additionally, we summarise the main claims of our work explicitly at the start of Section 2}.
  \item Did you describe the limitations of your work?
    \answerYes{} \textit{We provide a limitations section at the end of the paper}
  \item Did you discuss any potential negative societal impacts of your work? \\
    \answerYes{} \textit{We provide a broader impact section following the conclusion of the paper}.
  \item Have you read the ethics review guidelines and ensured that your paper conforms to them? \\
    \answerYes{}
\end{enumerate}

\item If you are including theoretical results...
\begin{enumerate}
  \item Did you state the full set of assumptions of all theoretical results? \\
    \answerYes{} \textit{The assumptions are clearly stated in the theorems} 
	\item Did you include complete proofs of all theoretical results? \\
    \answerYes{} \textit{All proofs are relegated into the supp matt}
\end{enumerate}

\item If you ran experiments...
\begin{enumerate}
  \item Did you include the code, data, and instructions needed to reproduce the main experimental results (either in the supplemental material or as a URL)? \\
    \answerYes{} \textit{We include code to implement our experiments in the supplementary material}.
  \item Did you specify all the training details (e.g., data splits, hyperparameters, how they were chosen)? \\
    \answerYes{} \textit{We provide a description of the data sets, splits and hyperparameters in the experiment section of the paper (the splits we followed are standard within the community).}
	\item Did you report error bars (e.g., with respect to the random seed after running experiments multiple times)?
    \answerYes{} \textit{We provide error bars for CIFAR-$100$ and CIFAR-$10$ but not for the ImageNet experiments which are significantly more computationally costly}
	\item Did you include the total amount of compute and the type of resources used (e.g., type of GPUs, internal cluster, or cloud provider)?
    \answerNo{} \textit{Whilst we used an internal University cluster and were at no point able to access more than $8$ $14GB$ NVIDIA-GPUs for a limited amount of time, since the total work spanned over a large period of time it is unreasonable to record the exact GPU hours for all the experiments. Although the total number of ImageNet runs will be in the order of tens for the full runs.}
\end{enumerate}

\item If you are using existing assets (e.g., code, data, models) or curating/releasing new assets...
\begin{enumerate}
  \item If your work uses existing assets, did you cite the creators? \\
    \answerYes{} \textit{We cite the datasets used in this work. For our software implementation, we reference the open-source libraries that we build on.}.
  \item Did you mention the license of the assets? \\
    \answerNo{} \textit{With respect to datasets, we make use of public domain datasets and cite the original works which in turn provide licensing information - however, we don't include an explicit list of licenses for each dataset (e.g. ImageNet~\citep{russakovsky2015imagenet}, etc.)}.
  \item Did you include any new assets either in the supplemental material or as a URL? \\
    \answerYes{} \textit{We include code for our experiments.}
  \item Did you discuss whether and how consent was obtained from people whose data you're using/curating? \\
    \answerNo{} \textit{In this work, we do not curate new data, but instead restrict our experiments to datasets that are widely used within the community (CIFAR-10, CIFAR-100, PTB and ImageNet) to allow comparison to prior work. We do not discuss the consent for these datasets.}
  \item Did you discuss whether the data you are using/curating contains personally identifiable information or offensive content? \\
    \answerNo{} \textit{For the datasets we use  (CIFAR-10, CIFAR-100, PTB and ImageNet), we do not discuss personally identifiable information or offensive content.}
\end{enumerate}

\item If you used crowdsourcing or conducted research with human subjects...
\begin{enumerate}
  \item Did you include the full text of instructions given to participants and screenshots, if applicable?
    \answerNA{}
  \item Did you describe any potential participant risks, with links to Institutional Review Board (IRB) approvals, if applicable?
    \answerNA{}
  \item Did you include the estimated hourly wage paid to participants and the total amount spent on participant compensation?
    \answerNA{}
\end{enumerate}

\end{enumerate}